\RequirePackage{fix-cm}
\documentclass[smallcondensed, natbib]{svjour3}
\smartqed  

\usepackage{graphicx, xcolor}

\usepackage{amsmath, amssymb}
\usepackage{mathtools}
\usepackage{physics}    

\usepackage{float}      
\usepackage{stfloats}   
\usepackage{caption}
\usepackage{subcaption}
\usepackage{adjustbox}
\usepackage{placeins}   
\usepackage{lineno}
\modulolinenumbers[5]

\usepackage{etoolbox}
\usepackage[algo2e, ruled, noend, commentsnumbered]{algorithm2e}

\newcommand{\ADTW}{ADTW}
\newcommand{\DTW}{DTW}

\newcommand{\PF}{PF}  
\newcommand{\PFvcfe}{\ensuremath{\mathit{PF}^+}}  

\newcommand{\DTWi}{DTW_{w=\infty}}

\newcommand{\WDTW}{\ensuremath{\mathit{WDTW}}}

\newcommand{\DDTW}{\ensuremath{\mathit{DDTW}}}
\newcommand{\DWDTW}{\ensuremath{\mathit{DWDTW}}}

\newcommand{\ERP}{\ensuremath{\mathit{ERP}}}
\newcommand{\MSM}{\ensuremath{\mathit{MSM}}}
\newcommand{\TWE}{\ensuremath{\mathit{TWE}}}
\newcommand{\LCSS}{\ensuremath{\mathit{LCSS}}}
\newcommand{\SQED}{SQED}

\newcommand{\NN}{\mathit{NN}}

\newcommand{\RRR}{\mathcal{R}}

\DeclareMathOperator{\cost}{\lambda}

\DeclareMathOperator{\tslen}{\ell}

\def\update#1#2{\textcolor{black}{#2}}
\def\newtext#1{\update{}{#1}}

\begin{document}


\title{Parameterizing the cost function of Dynamic Time Warping with application to time series classification%
\thanks{This work was supported by the Australian Research Council award DP210100072.}}
\author{Matthieu Herrmann \and Chang Wei Tan \and Geoffrey I. Webb }
\institute{Matthieu Herrmann \and Chang Wei Tan \and Geoffre I Webb. \at
    Monash University, Clayton Campus      \\
    Woodside Building, 20 Exhibition Walk  \\
    Monash University VIC 3800, Australia  \\
    \email{matthieu.herrmann@monash.edu}   \\      
    \email{chang.tan@monash.edu}           \\      
    \email{geoff.webb@monash.edu}          \\      
}

\date{Received: date / Accepted: date}

\maketitle

\begin{abstract}
Dynamic Time Warping ($\DTW$) is a popular time series distance measure
that aligns the points in two series with one another.
These alignments support warping of the time dimension to allow for processes that unfold at differing rates.
The distance is the minimum sum of costs of the resulting alignments
over any allowable warping of the time dimension.
The cost of an alignment of two points is a function of the difference in the values of those points.
The original cost function was the absolute value of this difference.
Other cost functions have been proposed. A popular alternative is the square of the difference.
However, to our knowledge, this is the first investigation of both the relative impacts of using different cost functions and the potential to tune cost functions to different time series classification tasks.
We do so in this paper by using a tunable cost function $\lambda_{\gamma}$ with parameter $\gamma$.
\newtext{We show that higher values of $\gamma$ place greater weight on larger pairwise differences, while lower values place greater weight on smaller pairwise differences.}
We~demonstrate that training $\gamma$ significantly improves the accuracy of both the $\DTW$ nearest neighbor
and Proximity Forest classifiers.
\keywords{Time Series\and Classification\and Dynamic Time Warping\and Elastic Distances}
\end{abstract}



\section{Introduction}

Similarity and distance measures are fundamental to data analytics,
supporting many key operations including 
similarity search \citep{rakthanmanon2012searching},
classification \citep{shifazTSCHIEFScalableAccurate2020},
regression \citep{tan2021regression},
clustering \citep{petitjean2011global},
anomaly and outlier detection \citep{diab2019anomaly},
motif discovery \citep{alaee2021time}, 
forecasting \citep{bandara2021improving},
and subspace projection \citep{DENG2020107210}.

Dynamic Time Warping ($\DTW$) \citep{sakoe1971, sakoe1978}
is a popular distance measure for time series and
is often employed as a similarity measure such that the lower the distance the greater the similarity.
It is used in numerous applications including speech recognition~\citep{sakoe1971, sakoe1978},
gesture recognition~\citep{cheng2016image},
signature verification~\citep{OKAWA2021107699},
shape matching~\citep{yasseen2016shape},
road surface monitoring~\citep{singh2017smart},
neuroscience~\citep{cao2016real}
and medical diagnosis~\citep{varatharajan2018wearable}.

$\DTW$ aligns the points in two series and returns the sum of the pairwise-distances
between each of the pairs of points in the alignment.
$\DTW$ provides flexibility in the alignments to allow for series that evolve at differing rates.
In the univariate case, pairwise-distances are usually calculated using a \emph{cost function},
$\lambda(a\in\RRR, b\in\RRR)\rightarrow \RRR^{+}$.
When introducing $\DTW$,
\cite{sakoe1971} defined the cost function as \update{the absolute value
of the difference between the aligned points, i.e.\ $\lambda(a,b) = \abs{a-b}$}{$\lambda(a,b) = \abs{a-b}$}.
However, other cost functions have subsequently been used.
The cost function
$\lambda(a,b) = (a-b)^2$~\citep{tan2018efficient,dauUCRTimeSeries2019,mueenExtractingOptimalPerformance2016,loningSktimeUnifiedInterface,tanFastEEFastEnsembles2020}~is now widely used,
possibly inspired by the (squared) Euclidean distance.
ShapeDTW~\citep{zhaoShapeDTWShapeDynamic2018} computes the cost between two points
by computing the cost between the ``shape descriptors'' of these points.
Such a descriptor can be the Euclidean distance between segments centered on this points,
taking into account their local neighborhood.

\begin{figure}
    \captionsetup[subfigure]{aboveskip=-2pt,belowskip=7pt}
    \centering
    \begin{subfigure}[b]{0.32\textwidth}
        \includegraphics[width=\textwidth]{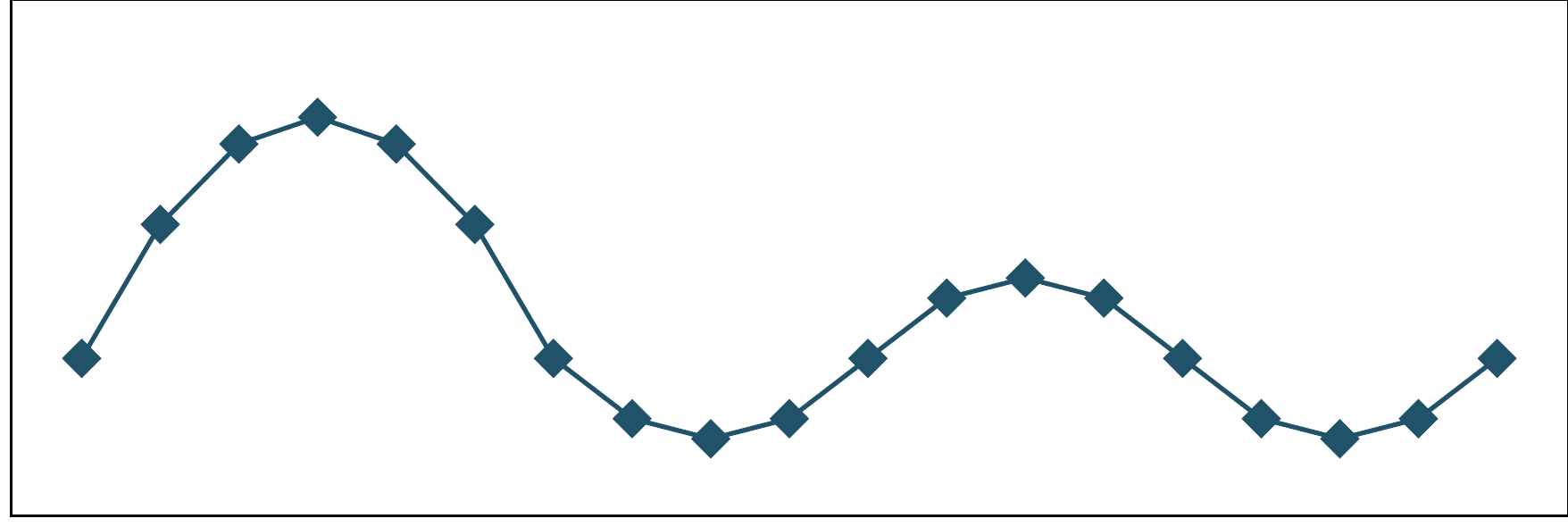}
        \vspace{0pt}
        \caption{\label{fig:hl_s}Series $S$}
    \end{subfigure}
    \hfill
    \begin{subfigure}[b]{0.32\textwidth}
        \includegraphics[width=\textwidth]{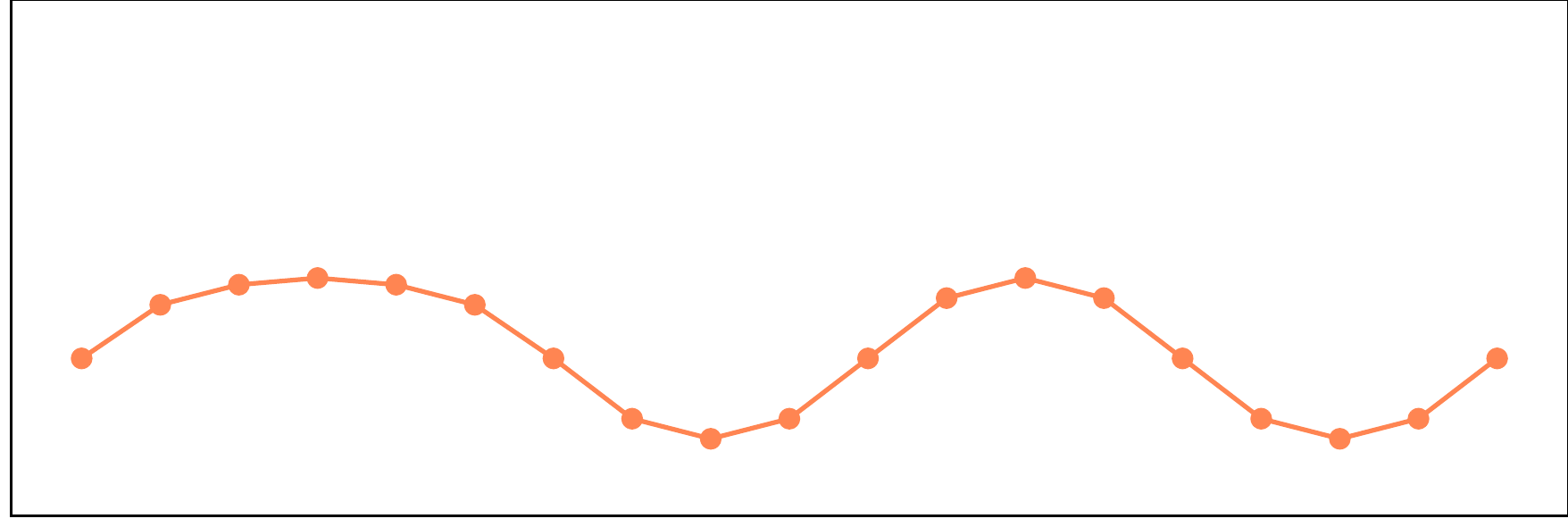}
        \vspace{0pt}        \caption{\label{fig:hl_t}Series $T$}
    \end{subfigure}
    \hfill
    \begin{subfigure}[b]{0.32\textwidth}
        \includegraphics[width=\textwidth]{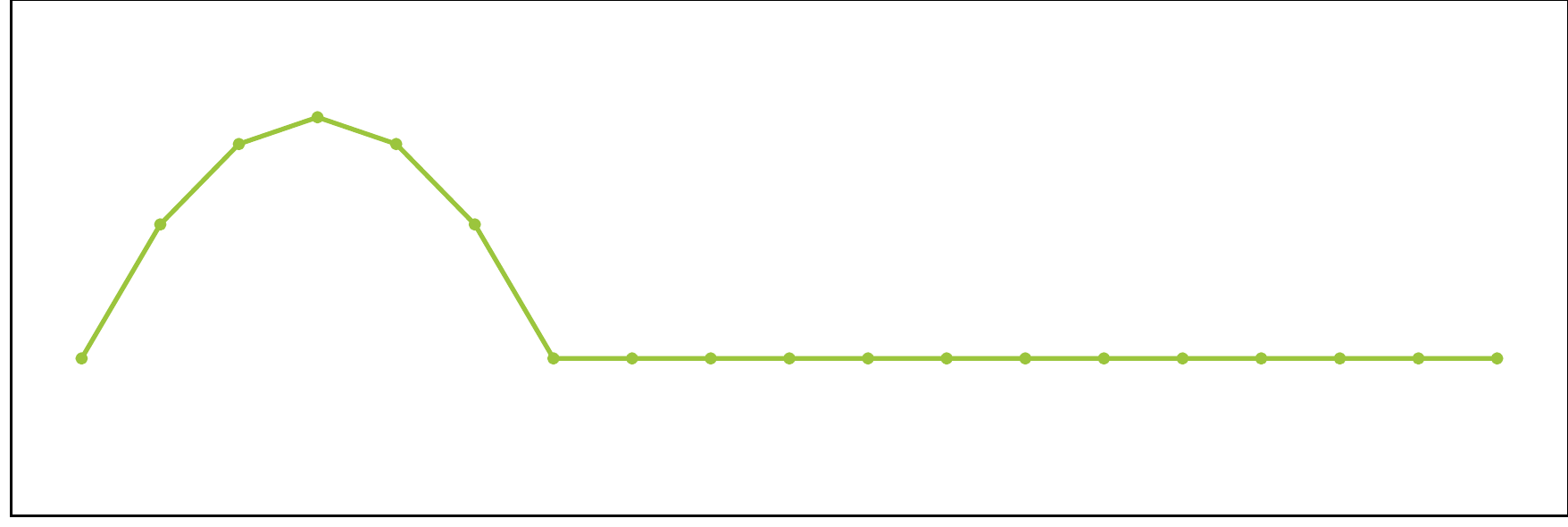}
        \vspace{0pt}        \caption{\label{fig:hl_u}Series $U$}
    \end{subfigure}
    \vspace*{-15pt}
    \caption{\label{fig:headline}
        Tuning the cost function changes which series are considered more similar to one another.
        $U$ exactly matches the first 7 points of $S$,
        but then flattens, running through the center of the remaining points in $S$.
        In contrast, $T$ starts with lower amplitude than $S$ over the first seven points,
        but then exactly matches $S$ for the remaining low amplitude waves.
        The original $\DTW$ cost function, $\cost(a, b)=\abs{a-b}$,
        results in $\DTW(S,T)=\DTW(S,U)=9$, with $\DTW$ rating $T$ and $U$ as equally similar to $S$.
        The commonly used cost function, $\cost(a, b)=(a-b)^2$,
        results in $\DTW(S,U)=9.18<\DTW(S,T)=16.66$.
        More weight is placed on the high amplitude start, and $S$ is more similar to $U$.
        Using the cost function $\cost(a, b)=\abs{a-b}^{0.5}$
        results in $\DTW(S,U)=8.98>\DTW(S,T)=6.64$,
        placing more weight on the low amplitude end, and $S$ is more similar to $T$.
        In general, changing the cost function alters the amount of weight placed
        on low amplitude vs high amplitude effects,
        allowing $\DTW$ to be better tuned to the varying needs of different applications.
        }
\end{figure}

To our knowledge, there has been little research into the influence of tuning the cost function
on the efficacy of $\DTW$ in practice.
This paper specifically investigates how actively tuning the cost function influences the outcome
on a clearly defined benchmark.
We do so using 
$\lambda_\gamma(a,b)=\abs{a-b}^\gamma$ as the cost function for $\DTW$,
where $\gamma=1$ gives us the original cost function;
and $\gamma=2$ the now commonly used squared Euclidean distance.

We motivate this research with an example illustrated in Figure~\ref{fig:headline}
relating to three series, $S$, $T$ and $U$.
$U$ exactly matches $S$ in the high amplitude effect at the start,
but does not match the low amplitude effects thereafter.
$T$ does not match the high amplitude effect
at the start but exactly matches the low amplitude effects thereafter.
Given these three series, we can ask which of $T$ or $U$ is the nearest neighbor of $S$? 

As shown in Figure~\ref{fig:headline}, the answer varies with~$\gamma$.
\newtext{Low $\gamma$ emphasizes low amplitude effects and hence identifies $S$ as more similar to $T$, while high $\gamma$ emphasizes high amplitude effects and assesses $U$ as most similar to $S$.}
Hence, we theorized that careful selection of an effective cost function
on a task by task basis can greatly improve accuracy,
which we demonstrate in a set of nearest neighbor time series classification experiments.
Our findings extend directly to all applications relying on nearest neighbor search,
such as ensemble classification
(we demonstrate this with Proximity Forest~\citep{lucasProximityForestEffective2019})
and clustering, and have implications for all applications of $\DTW$.

The remainder of this paper is organized as follows.
In Section~\ref{sec:background}, we provide a detailed introduction to $\DTW$ and its variants.
In Section~\ref{sec:cost-func}, we present the flexible parametric cost function $\lambda_\gamma$
and a straightforward method for tuning its parameter.
Section~\ref{sec:experiments} presents experimental assessment of the impact
of different $\DTW$ cost functions,
and the efficacy of $\DTW$ cost function tuning in \update{nearest neighbor classification}{similarity-based time series classification (TSC)}.
Section~\ref{sec:conclusion} provides discussion, directions for future research and conclusions.

\section{Background}\label{sec:background}

\subsection{Dynamic Time Warping}
\label{subsec:dtw}
\update{The $\DTW$ distance was first introduced in 1971 for aligning and comparing time series
with application to speech recognition \citep{sakoe1971}.
It has been one of the most widely used distance measures}{The $DTW$ distance measure \citep{sakoe1971} is widely used}  in many time series data analysis tasks,
\update{such as}{including} nearest neighbor ($\NN$) search
\citep{rakthanmanon2012searching,tan2021regression,petitjean2011global,keoghDerivativeDynamicTime2001,silva2018}.
Nearest neighbor with $\DTW$ ($\NN$-$\DTW$) has been the historical approach to time series classification
and is still used widely today.

\begin{figure}
    \hfil\includegraphics[width=0.9\columnwidth]{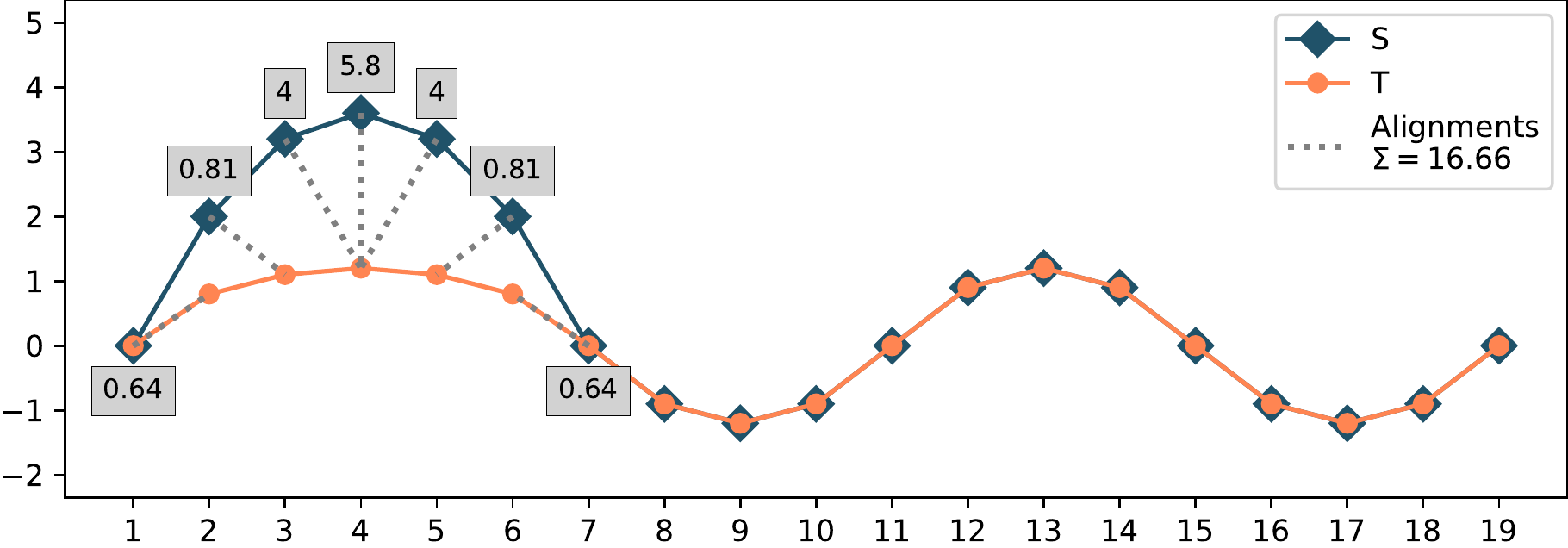}\hfil
    \caption{\label{fig:DTW:Alignment}
        Pairwise alignments of $\DTW(S,T)$ with $\gamma=2$, accumulating a total cost of $16.66$.
        We only show non-zero alignments.}
\end{figure}

$\DTW$ computes the cost of an optimal alignment between two equal length series,
$S$ and $T$ with length $L$ in $O(L^2)$ time (lower costs indicating more similar series),
by minimizing the cumulative cost of aligning their individual points, also known as the warping path.
The warping path of $S$ and $T$ is a sequence $\mathcal{W}=\langle\mathcal{W}_1,\ldots,\mathcal{W}_P\rangle$
of alignments (dotted lines in Figure~\ref{fig:DTW:Alignment}).
Each alignment is a pair $\mathcal{W}_k=(i,j)$ indicating that $S_i$ is aligned with $T_j$. 
$\mathcal{W}$ must obey the following constraints: 
\begin{itemize}
  \item {\bf Boundary Conditions}: $\mathcal{W}_1=(1,1)$ and $\mathcal{W}_P=(L,L)$.
  \item {\bf Continuity} and {\bf Monotonicity}:
    for any $\mathcal{W}_k=(i,j)$, $1<k\leq{}P$, we have
    $\mathcal{W}_{k+1}\in\{(i{+}1,j), (i,j{+}1), (i{+}1,j{+}1)\}$.
\end{itemize}

\begin{figure}
  \centering
  \subfloat[\label{fig:DTW:cm1}$\DTW^{1}(S,T)$ with $w=2$ ]{%
    \includegraphics[trim=0 0 0 0, clip,width=0.48\linewidth]{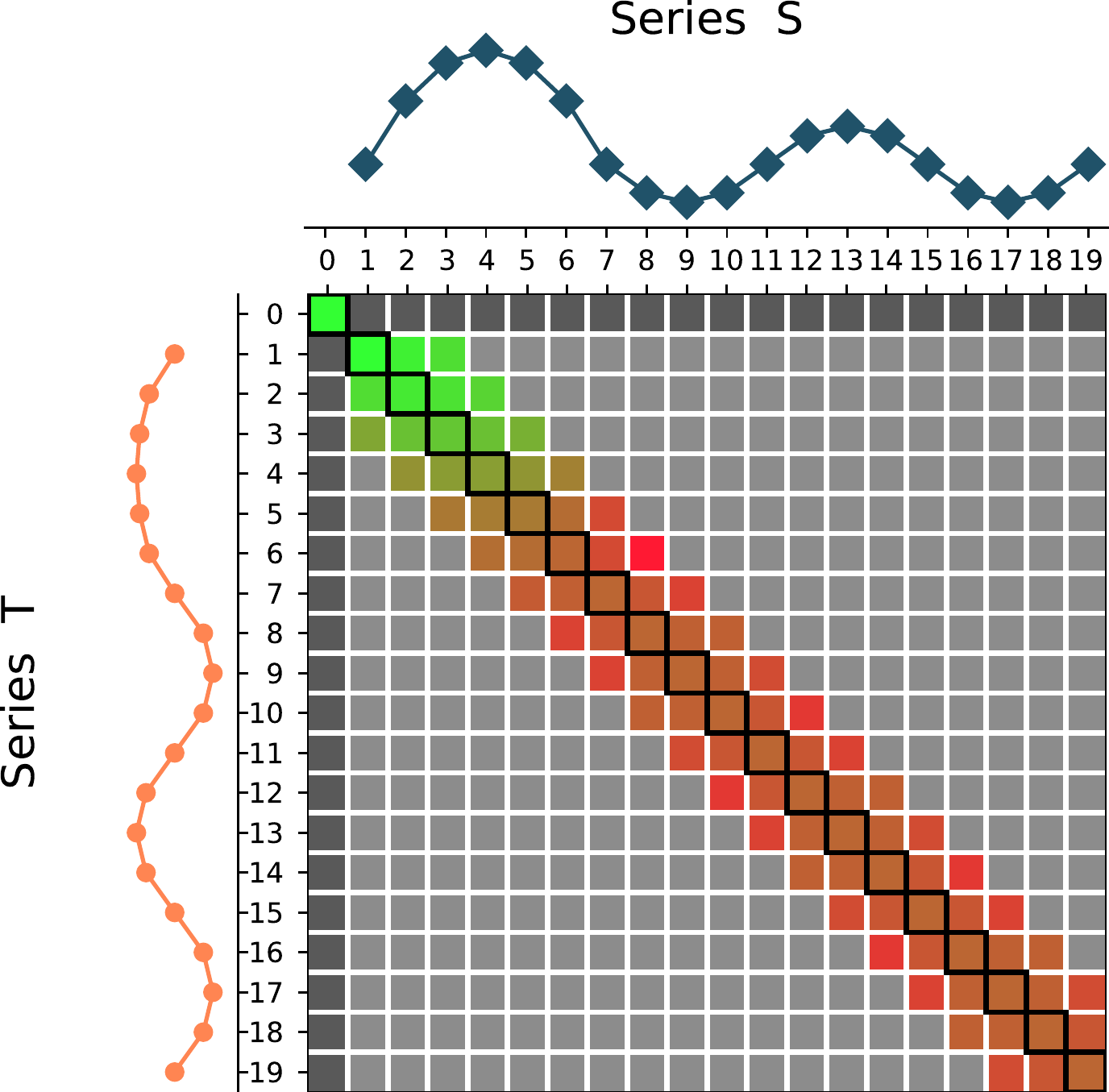}%
  }\hfill%
  \subfloat[\label{fig:DTW:cm2}$\DTW^{2}(S,T)$ with $w=2$]{%
    \includegraphics[trim=0 0 0 0, clip,width=0.48\linewidth]{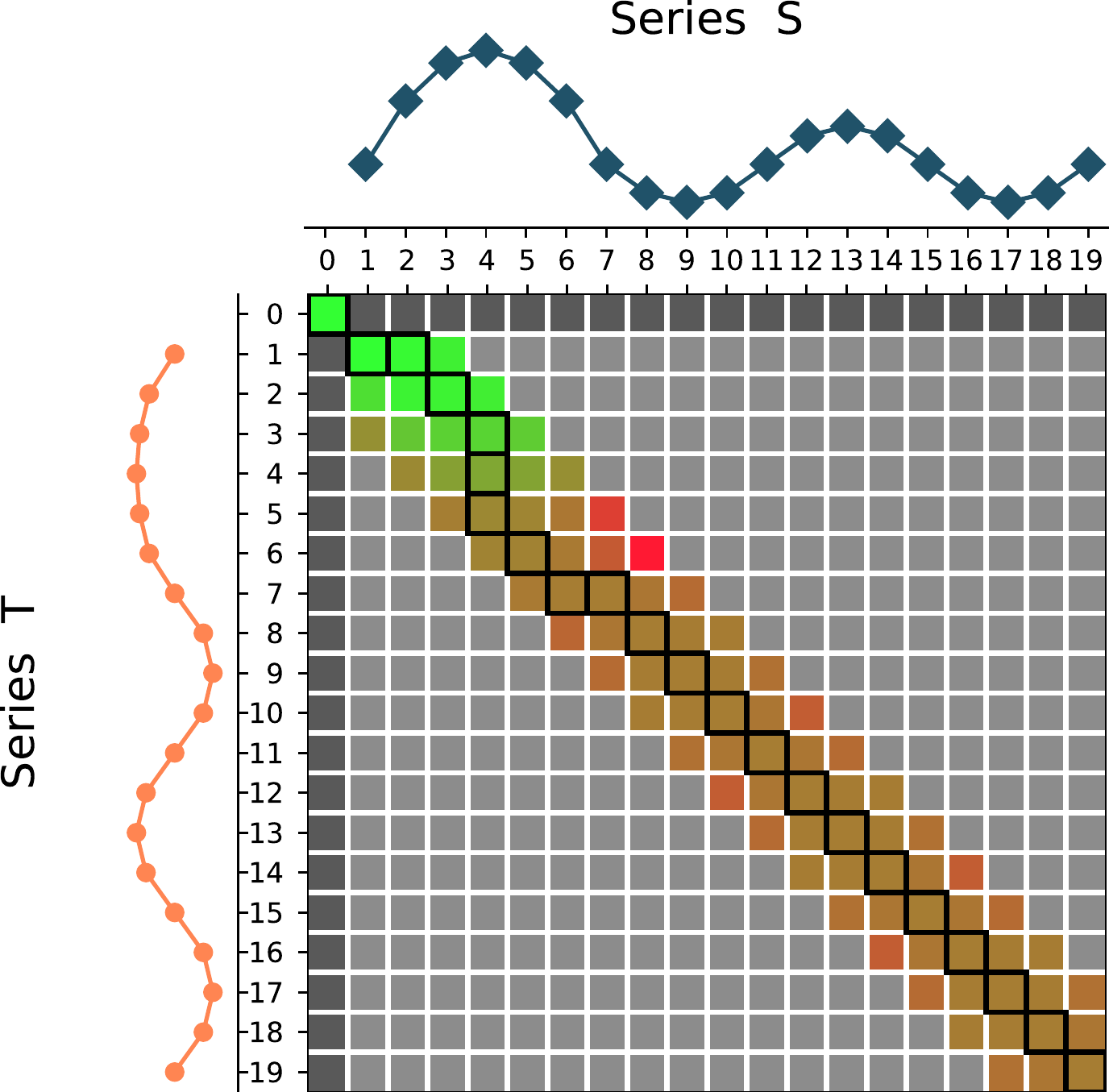}%
  }%
  \caption{\label{fig:DTW:cm}%
  $M_{\DTW(S,T)}$ with warping window $w=2$,
  and different cost function exponent,
  (a) $\gamma=1$ and (b) $\gamma=2$.
  We have $\DTW(S,T){=}M_{\DTW(S,T)}(L,L)$.
  The amplitude of the cumulative cost is represented by a green (minimal) to red (maximal) gradient.
  Cells cut-out by the warping window are in light gray, borders are in dark gray.
  The warping path cells are highlighted with black borders.
  Notice how the deviation from the diagonal in (b) corresponds to the alignments
  Figure~\ref{fig:DTW:Alignment}.
  }
\end{figure}

The cost of a warping path is minimized using dynamic programming
by building a ``cost matrix'' $M_{\DTW}$ for the two series $S$ and $T$,
such that $M_{\DTW}(i,j)$ is the minimal cumulative
cost of aligning the first $i$ points of $S$ with the first $j$ points of $T$.
The cost matrix is defined in Equations~\ref{eq:DTW:corner} to~\ref{eq:DTW:main},
where $\cost(S_i,T_j)$ is the cost of aligning the two points, discussed in Section \ref{sec:cost-func}.
It follows that $\DTW(S,T) {=} M_{\DTW}(L,L)$.

\begin{subequations}\label{eq:DTW}
\begin{align}
    M_{\DTW}(0,0) &= 0       \label{eq:DTW:corner}\\
    M_{\DTW}(i,0) &= +\infty \label{eq:DTW:vborder}\\ M_{\DTW}(0,j) &= +\infty \label{eq:DTW:hborder}\\
    M_{\DTW}(i,j) &= \cost(S_i, T_j) + \min\left\{
    \begin{aligned}
        &M_{\DTW}(i{-}1, j{-}1) \\
        &M_{\DTW}(i{-}1, j) \\
        &M_{\DTW}(i, j{-}1)
    \end{aligned}
    \right. \label{eq:DTW:main}
\end{align}
\end{subequations}

Figure~\ref{fig:DTW:cm} shows the cost matrix of computing $\DTW(S,T)$.
\update{The warping path is highlighted using the bold boxes going through the matrix. 
See how the vertical section of the path column 4 in Fig.~\ref{fig:DTW:cm2}
corresponds to the fourth point of $S$ being aligned thrice
in Fig.~\ref{fig:DTW:Alignment}.}{The warping path is highlighted using the bold boxes going through the matrix.}

$\DTW$ is commonly used with a global constraint applied on the warping path,
such that $S_i$ and $T_j$ can only be aligned if they are within a window range, $w$.
This limits the distance in the time dimension that can separate $S_i$
from points in $T$ with which it can be aligned \citep{sakoe1971,keogh2005exact}.  
This constraint is known as the warping window,
$w$ (previously Sakoe-Chiba band) \citep{sakoe1971}.
Note that we have $0\leq w\leq L-2$; 
$\DTW$ with $w=0$ corresponds to a \emph{direct alignment} in which $\forall_{(i,j)\in\mathcal{W}}\,i=j$;
and $\DTW$ with $w\update{=}{\geq}L-2$ places no constraints on the distance between the points in an alignment. 
\update{$L-2$ is the longest effective window as the cell at $(1, L)$ cannot be part of a warping path
because taking the diagonal from $(1, L-1)$ to $(2, L)$ is never more expensive than the steps
$(1, L-1)$ to $(1, L)$ followed by $(1, L)$ to $(2, L)$.
Figure~\ref{fig:DTW:cm} shows an example with warping window $w{=}2$,
where the alignment of $S$ and $T$ is constrained to be inside the colored band.}{Figure~\ref{fig:DTW:cm} shows an example with warping window $w{=}2$,
where the alignment of $S$ and $T$ is constrained to be inside the colored band.}
Light gray cells are ``forbidden'' by the window.

\update{Using a warping window has two main benefits:
(1) increasing $\NN$-$\DTW$ accuracy by preventing pathological warping of $S$ and $T$.
(2) speeding up $\NN$-$\DTW$ by reducing the complexity
    of $\DTW$ from $O(L^2)$ to $O(W\cdot L)$ \citep{tan2018efficient,tan2021ultra}.
}{Warping windows provide two main benefits:
(1) preventing pathological warping of $S$ and $T$; and 
(2) speeding up $\DTW$ by reducing its complexity from $O(L^2)$ to $O(W\cdot L)$ \citep{tan2018efficient,tan2021ultra}.}

Alternative window constraints have also been developed,
such as the Itakura Parallelogram \citep{itakuraMinimumPredictionResidual1975}
and the Ratanamahatana-Keogh band \citep{ratanamahatana2004making}.
In this paper, we focus on the Sakoe-Chiba Band
which is the constraint defined in the original definition of $\DTW$. 

\subsection{Amerced Dynamic Time Warping}
\label{subsec:adtw}
$\DTW$ uses a crude step function to constrain the alignments,
where any warping is allowed within the warping window and none beyond it.  
This is unintuitive for many applications,
where some flexibility in the exact amount of warping might be desired.
The Amerced Dynamic Time Warping ($\ADTW$) distance measure
is an intuitive and effective variant of $\DTW$ \citep{adtw}.
Rather than using a tunable hard constraint like the warping window,
it applies a tunable additive penalty $\omega$ for non-diagonal (warping) alignments \citep{adtw}. 

$\ADTW$ is computed with dynamic programming, similar to $\DTW$,
using a cost matrix $M_{\ADTW}$ with $\ADTW_{\omega}(S,T)=M_{\ADTW}(L,L)$.
Equations \ref{eq:ADTW:corner} to~\ref{eq:ADTW:main} describe this cost matrix,
where $\cost(S_i,T_j)$ is the cost of aligning the two points, discussed in Section \ref{sec:cost-func}.

\begin{subequations}\label{eq:ADTW}
\begin{align}
    M_{\ADTW}(0,0) &= 0       \label{eq:ADTW:corner}\\
    M_{\ADTW}(i,0) &= +\infty \label{eq:ADTW:vborder}\\
    M_{\ADTW}(0,j) &= +\infty \label{eq:ADTW:hborder}\\
    M_{\ADTW}(i,j) &= \min\left\{
    \begin{aligned}
        &M_{\ADTW}(i{-}1, j{-}1) + \cost(S_i, T_j)\\
        &M_{\ADTW}(i{-}1, j) + \cost(S_i, T_j) + \omega \\
        &M_{\ADTW}(i, j{-}1) + \cost(S_i, T_j) + \omega
    \end{aligned}
    \right. \label{eq:ADTW:main}
\end{align}
\end{subequations}

The parameter $\omega$ works similarly to the warping window,
allowing $\ADTW$ to be as flexible as $\DTW$ with $w=L-2$, and as constrained as $\DTW$ with $w=0$.
A small penalty should be used if large warping is desirable, while large penalty minimizes warping.
Since $\omega$ is an additive penalty, its scale relative to the time series in context matters,
as a small penalty in a given problem maybe a huge penalty in another one.
An automated parameter selection method has been proposed in the context of time series classification
that considers the scale of $\omega$ \citep{adtw}.
The scale of penalties is determined by multiplying the maximum penalty
$\omega'$ by a ratio $0 \leq r \leq 1$, i.e. $\omega=\omega'\times r$.
The maximum penalty $\omega'$ is set to the average ``direct alignment''
sampled randomly from pairs of series in the training dataset, using the specifed cost function.
A direct alignment does not allow any warping, and corresponds to the diagonal of the cost matrix
(e.g. the warping path in Figure~\ref{fig:DTW:cm1}).
Then 100 ratios are sample from $r_i=(\frac{i}{100})^5$ for $1\leq i\leq100$
to form the search space for $\omega$.
Apart from being more intuitive, $\ADTW$ when used in a $\NN$ classifier
is significantly more accurate than $\DTW$ on 112 UCR time series benchmark datasets~\citep{adtw}.

\newtext{Note that $\omega$ can be considered as a direct penalty on path length. If series $S$ and $T$ have length $L$ and the length of the warping path for $\ADTW_\omega(S,T)$ is $P$, the sum of the $\omega$ terms added will equal $2\omega(P-L+1)$. The longer the path, the greater the penalty added by $\omega$.}
\section{Tuning the cost function}\label{sec:cost-func}

$\DTW$ was originally introduced with the cost function $\lambda(a,b) = \abs{a-b}$.
Nowadays, the cost function $\lambda(a,b)=(a-b)^2=\abs{a-b}^2$ is also widely used~%
\citep{dauUCRTimeSeries2019,mueenExtractingOptimalPerformance2016,loningSktimeUnifiedInterface,tanFastEEFastEnsembles2020}. Some generalizations of $\DTW$ have also included tunable cost functions \citep{Deriso2022}.
To~our knowledge, the relative strengths and weaknesses of these two common cost functions
has not previously been thoroughly evaluated.
To study the impact of the cost function on $\DTW$, and its recent refinement $\ADTW$,
we use the cost function $\lambda_{\gamma}(a,b)=\abs{a-b}^{\gamma}$.

\begin{figure}
    \centering
    \includegraphics[width=0.5\textwidth]{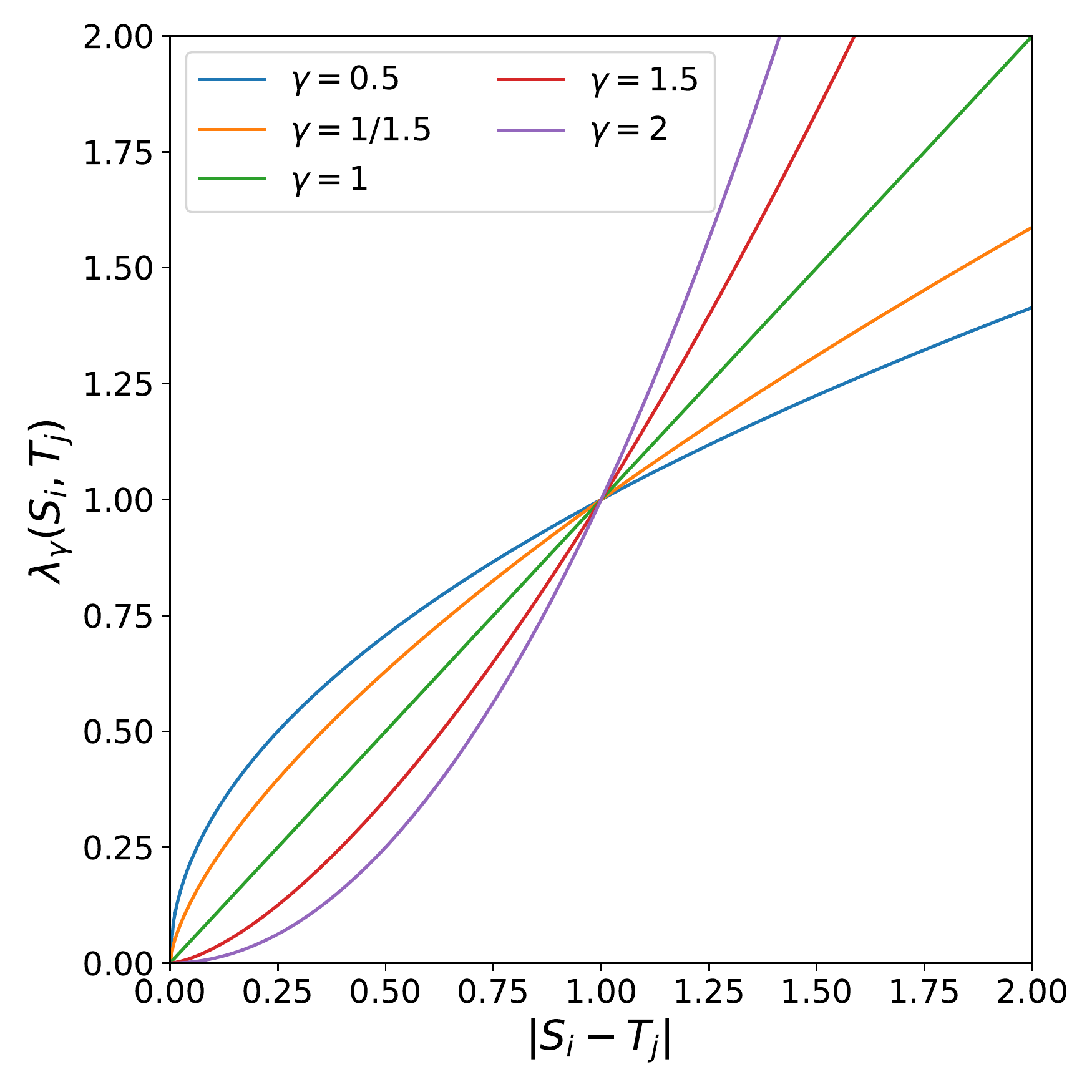}
    \caption{\label{fig:gammafam}
        Illustration of the effect of $\gamma\in\{1/2, 1/1.5, 1, 1.5, 2\}$ on $\lambda_\gamma$.}
\end{figure}

We primarily study the cost functions $\lambda_{\gamma}$
for $\gamma\in\Gamma=\{1/2, 1/1.5, 1, 1.5, 2\}$.
This includes the original $\DTW$ cost function $\abs{a-b}=\lambda_{1}(a,b)$,
and the more recent $(a-b)^2=\lambda_{2}(a,b)$.
To the best of our knowledge, the remaining cost functions, \update{such as
$\lambda_{0.5}(a,b)=\abs{a-b}^{0.5}$}{$\lambda_{0.5}(a,b)$, $\lambda_{0.\dot{6}}(a,b)$ and $\lambda_{1.5}(a,b)$}, have not been  previously investigated.
As~illustrated in Figure~\ref{fig:gammafam},
Relative to 1,
larger values of $\gamma$ penalize small differences less, and larger differences more.
Reciprocally, smaller values of $\gamma$ penalize large differences more,
and small differences less.

We will show in Section~\ref{sec:experiments} that learning $\gamma$ at train time over these
5 values is already enough to significantly improve nearest neighbor classification test accuracy.
We will also show that expanding $\Gamma$ to
a \emph{larger} set $\{1/5, 1/4, 1/3, 1/2, 1/1.5,\allowbreak 1, 1.5, 2, 3, 4, 5\}$,
or a \emph{denser} set $\{1/2, 1/1.75, 1/1.5, 1/1.25, 1, 1.25, 1.5, 1.75, 2\}$
does not significantly improve the classification accuracy,
even with doubling the number of explored parameters.
Note that all the sets have the form $\{\frac{1}{n}\dots{}1\dots{}n\}$.
Although this balancing is not necessary, we did so to strike a balance in the available exponents.

Tuning $\lambda_{\gamma}$ amounts to learning the parameter $\gamma$ at train time.
This means that we now have two parameters for both $\DTW$
(the warping window $w$ and $\gamma$) and $\ADTW$ (the penalty $\omega$ and $\gamma$).
In the current work, the $w$ and $\omega$ parameters are always learned independently for each $\gamma$,
using the standard method~\citep{adtw}.
We denote $\DTW$ with $\lambda_x = |S_i-T_j|^x$ as $\DTW^{x}$,
and $\ADTW$ with $\lambda_x$ as $\ADTW^{x}$.
We indicate that the cost function has been tuned with the superscript $+$,
i.e. $\DTW^+$ and $\ADTW^+$.

Note that with a window $w=0$,
\begin{equation}
    \DTW^+(S, T) = \sum\limits_{i=1}^{n}{(S_i-T_i)^\gamma}
\end{equation}
for the selected exponent $\gamma$.
In other words, it is the Minkowski distance \citep{thompson1996minkowski} to the power $\gamma$,
providing the same relative order as the Minkowski distance,
i.e. they both have the same effect for nearest neighbor search applications.

The parameters $w$ and $\omega$ have traditionally been learned through
leave-one-out cross-validation (LOOCV) evaluating 100 parameter 
values~\citep{tan2018efficient,tanFastEEFastEnsembles2020,lines2015,tan2021ultra}.
Following this approach, we evaluate 100 parameter values for $w$ (and $\omega$)
per value of $\gamma$, i.e. we evaluate 500 parameter values for $\DTW^+$ and $\ADTW^+$.
To enable a fair comparison in Section~\ref{sec:experiments} of $\DTW^+$ (resp. $\ADTW^+$)
against $\DTW^\gamma$ (resp. $\ADTW^\gamma$) with fixed $\gamma$,
the latter are trained evaluating both 100 parameter values (to give the same space of values for $w$ or $\omega$)
as well as 500 parameter values (to give the same overall number of parameter values).

Given a fixed $\gamma$,
LOOCV can result in multiple parameterizations for which the train accuracy is equally best.
We need a procedure to break ties.
This could be achieved through random choice, in which case the outcome becomes nondeterministic (which may be desired).
Another possibility is to pick a parameterization depending on other considerations.
For $\DTW$, we pick the smallest windows as it leads to faster computations.
For $\ADTW$, we follow the paper~\citep{adtw} and pick the median value.

We also need a procedure to break ties when more than one pair of values
over two different parameters all achieve equivalent best performance.
We do so by forming a hierarchy over the parameters.
We first pick a best value for $w$ (or $\omega$) per possible $\gamma$,
forming dependent pairs $(\gamma, w)$ (or $(\gamma, \omega)$).
Then, we break ties between pairs
by picking the one with the median $\gamma$.
In case of an even number of equal best values for $\gamma$,
taking a median would result in taking an average of dependent pairs,
which does not make sense for the dependent value ($w$ or $\omega$).
In this case we select between the two \emph{middle} pairs the one with a $\gamma$ value closer to $1$,
biasing the system towards a balanced response to differences less than or greater than zero.

Our method does not change the overall time complexity of learning $\DTW$'s and $\ADTW$'s parameters.
The time complexity of using LOOCV for nearest neighbor search with this distances is $O(M.N^2.L^2)$,
where $M$ is the number of parameters, $N$ is the number of training instances, and $L$ is the length of the series.
Our method only impacts the number of parameters $M$.
Hence, using 5 different exponents while keeping a hundred parameters for $w$ or $\omega$ effectively increases the training time 5 fold.

\section{Experimentation}\label{sec:experiments}


We evaluate the practical utility of cost function tuning by studying
its performance in nearest neighbor classification.
While the technique has potential applications well beyond classification,
we choose this specific application because it has well accepted benchmark problems
with objective evaluation criteria (classification accuracy).
We experimented over the widely-used time series classification benchmark
of the UCR archive~\citep{UCRArchive2018},
removing the datasets containing series of variable length
or classes with only one training exemplar, leading to 109 datasets. 
We~investigate tuning the exponent $\gamma$ for $\DTW^+$ and $\ADTW^+$ using the following sets
(and we write e.g. $\DTW^{+a}$ when using the set $a$):
\begin{itemize}
    \item The \emph{default} set $a = \{1/2, 1/1.5, 1, 1.5, 2\}$
    \item The \emph{large} set $b = \{1/5, 1/4, 1/3, 1/2, 1/1.5, 1, 1.5, 2, 3, 4, 5\}$
    \item The \emph{dense} set $c = \{1/2, 1/1.75, 1/1.5, 1/1.25, 1, 1.25, 1.5, 1.75, 2\}$
\end{itemize}
The default set $a$ is the one used
in Figure \ref{fig:gammafam},
and the one we recommend.

\newtext{We show that a wide range of different exponents $\gamma$ each perform best on different datasets.
We then compare $\DTW^{+a}$ and $\ADTW^{+a}$ against their classic counterparts using $\gamma=1$ and $\gamma=2$.
We also address the question of the number of evaluated parameters,
showing with both $\DTW$ and $\ADTW$ that tuning the cost function is more beneficial
than evaluating 500 values of either $w$ or $\omega$ with a fixed cost function.
We then show that compared to the large set $b$ (which looks at exponents beyond $1/2$ and $2$)
and to the dense set $c$ (which looks at more exponents between $1/2$ and $2$),
$a$ offer similar accuracy while being less computationally demanding
(evaluating less parameters).
Just as $\ADTW$ is significantly more accurate than $\DTW$ \citep{adtw},
$\ADTW^{+a}$ remains significantly more accurate than $\DTW^{+a}$. This holds for sets $b$ and $c$.}

Finally, we show that parameterizing the cost function is also beneficial in an ensemble classifier,
showing a significant improvement in accuracy for \newtext{ the leading similarity-based TSC algorithm,} Proximity Forest~\citep{lucasProximityForestEffective2019}.

\subsection{Analysis of the impact of exponent selection on accuracy}
\begin{figure}
    \captionsetup[subfigure]{aboveskip=-2pt,belowskip=7pt}
    \centering
    \begin{subfigure}[b]{0.49\textwidth}
        \includegraphics[width=\textwidth]{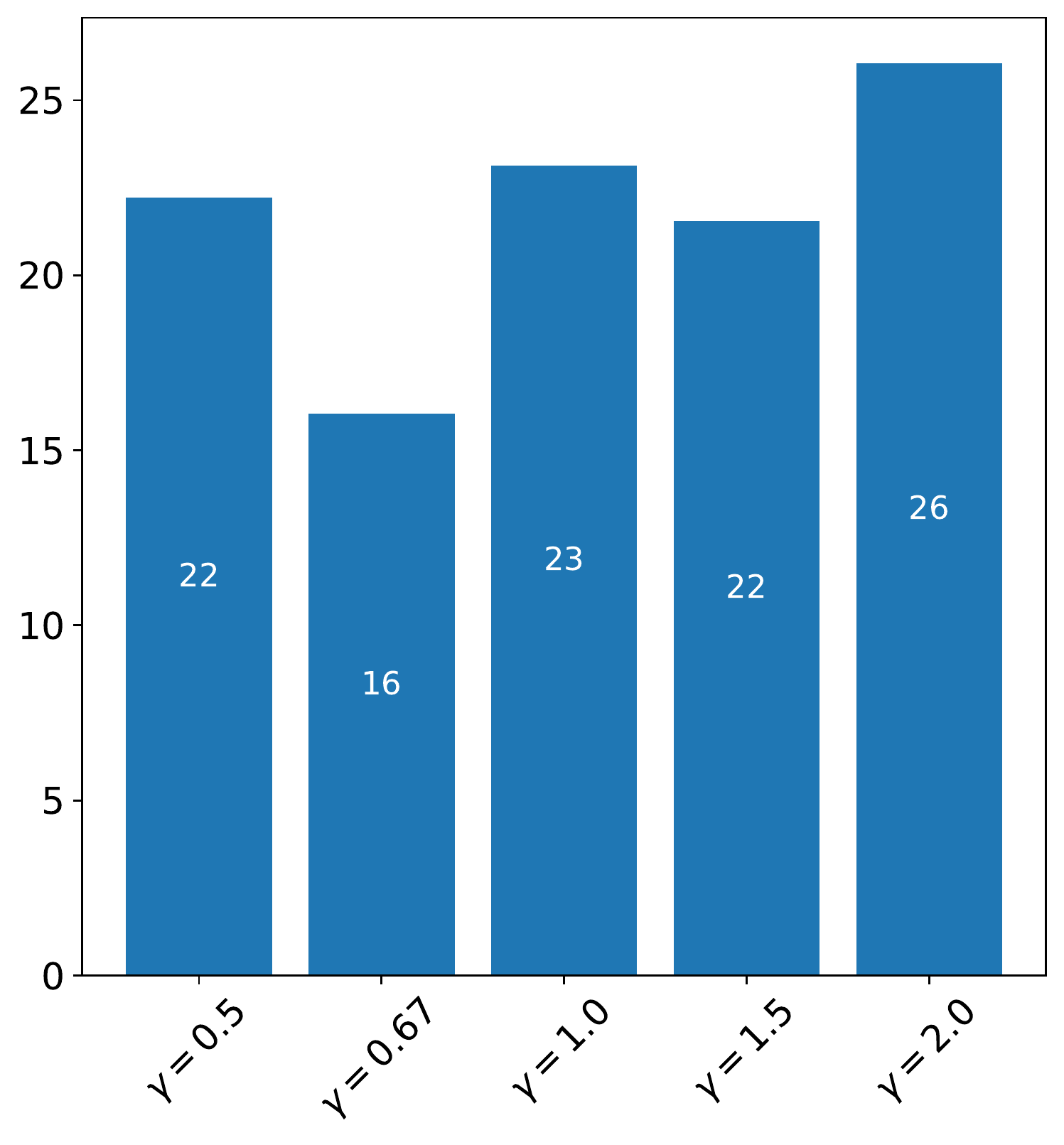}
        \caption{\label{fig:bplot:dtw_a}Most accurate $\DTW^{\gamma}$ for $\gamma\in a$}
    \end{subfigure}
    \hfill
    \begin{subfigure}[b]{0.49\textwidth}
        \includegraphics[width=\textwidth]{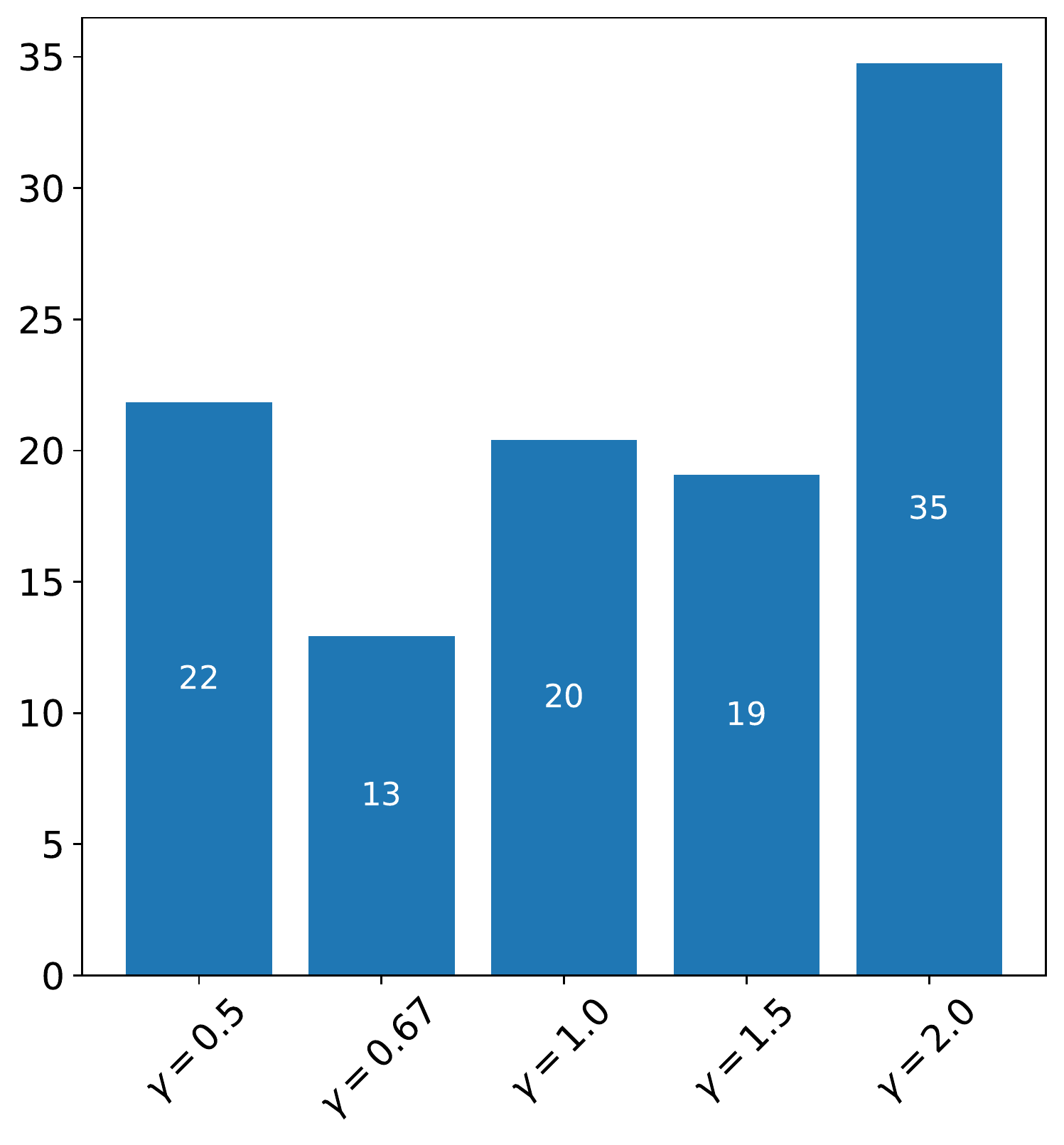}
        \caption{\label{fig:bplot:adtw_a}Most accurate $\ADTW^{\gamma}$ for $\gamma\in a$ }
    \end{subfigure}
    \\
    \centering
    \begin{subfigure}[b]{0.49\textwidth}
        \includegraphics[width=\textwidth]{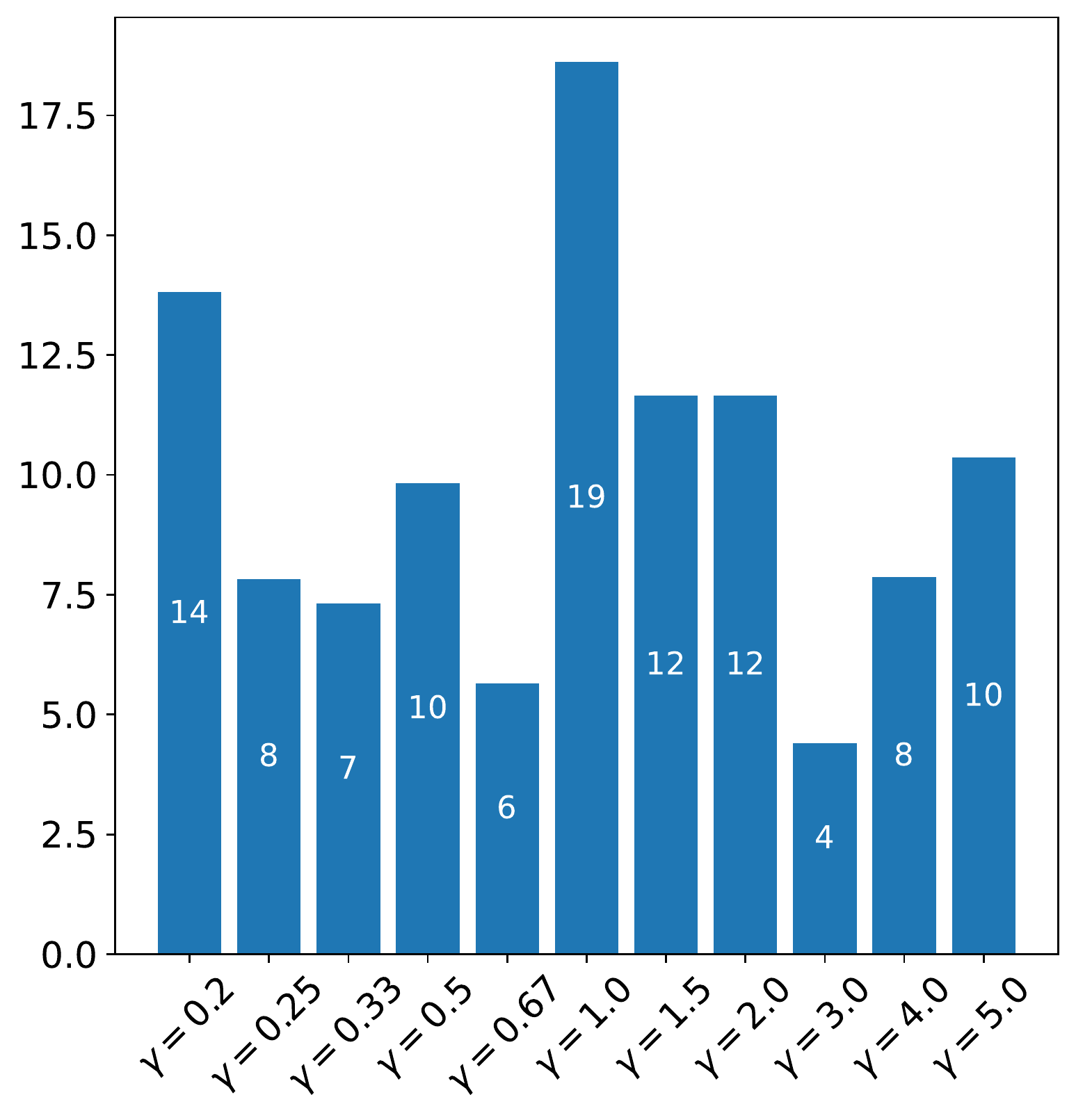}
        \caption{\label{fig:bplot:dtw_ab}Most accurate $\DTW^{\gamma}$ for $\gamma\in b$}
    \end{subfigure}
    \hfill
    \begin{subfigure}[b]{0.49\textwidth}
        \includegraphics[width=\textwidth]{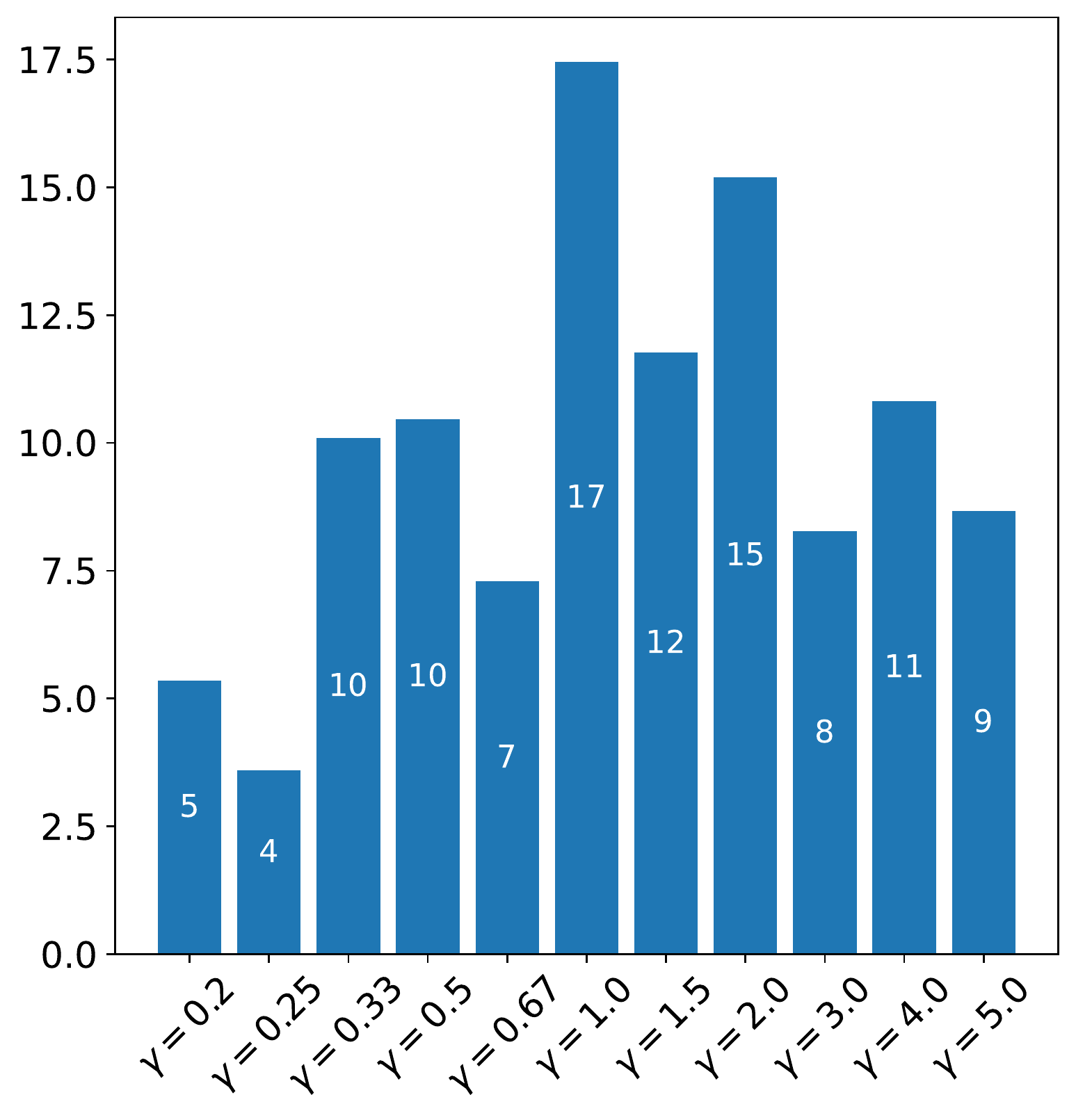}
        \caption{\label{fig:bplot:adtw_ab}Most accurate $\ADTW^{\gamma}$ for $\gamma\in b$}
    \end{subfigure}
    \\
    \centering
    \begin{subfigure}[b]{0.49\textwidth}
        \includegraphics[width=\textwidth]{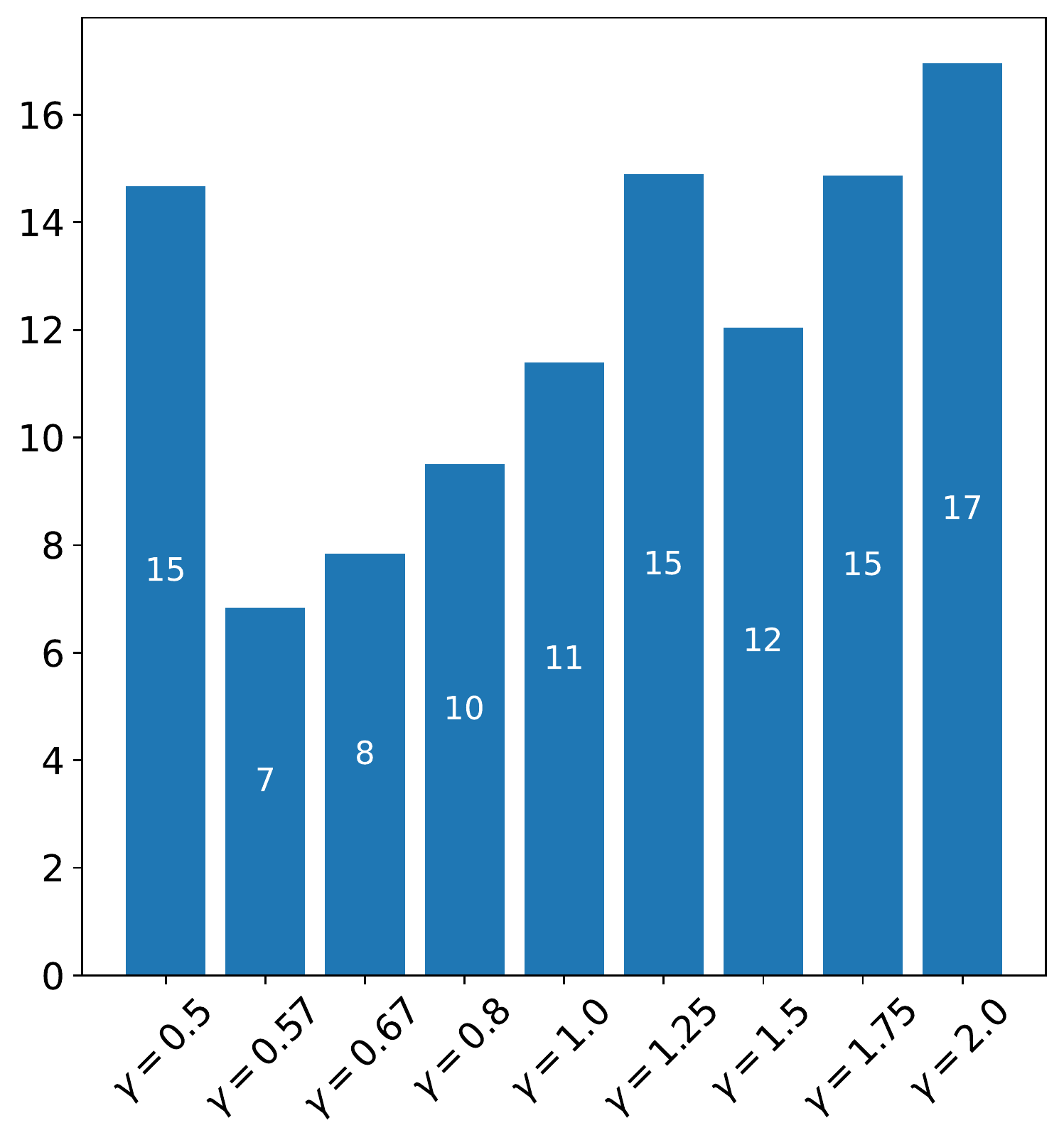}
        \caption{\label{fig:bplot:dtw_ac}Most accurate $\DTW^{\gamma}$ for $\gamma\in c$}
    \end{subfigure}
    \hfill
    \begin{subfigure}[b]{0.49\textwidth}
        \includegraphics[width=\textwidth]{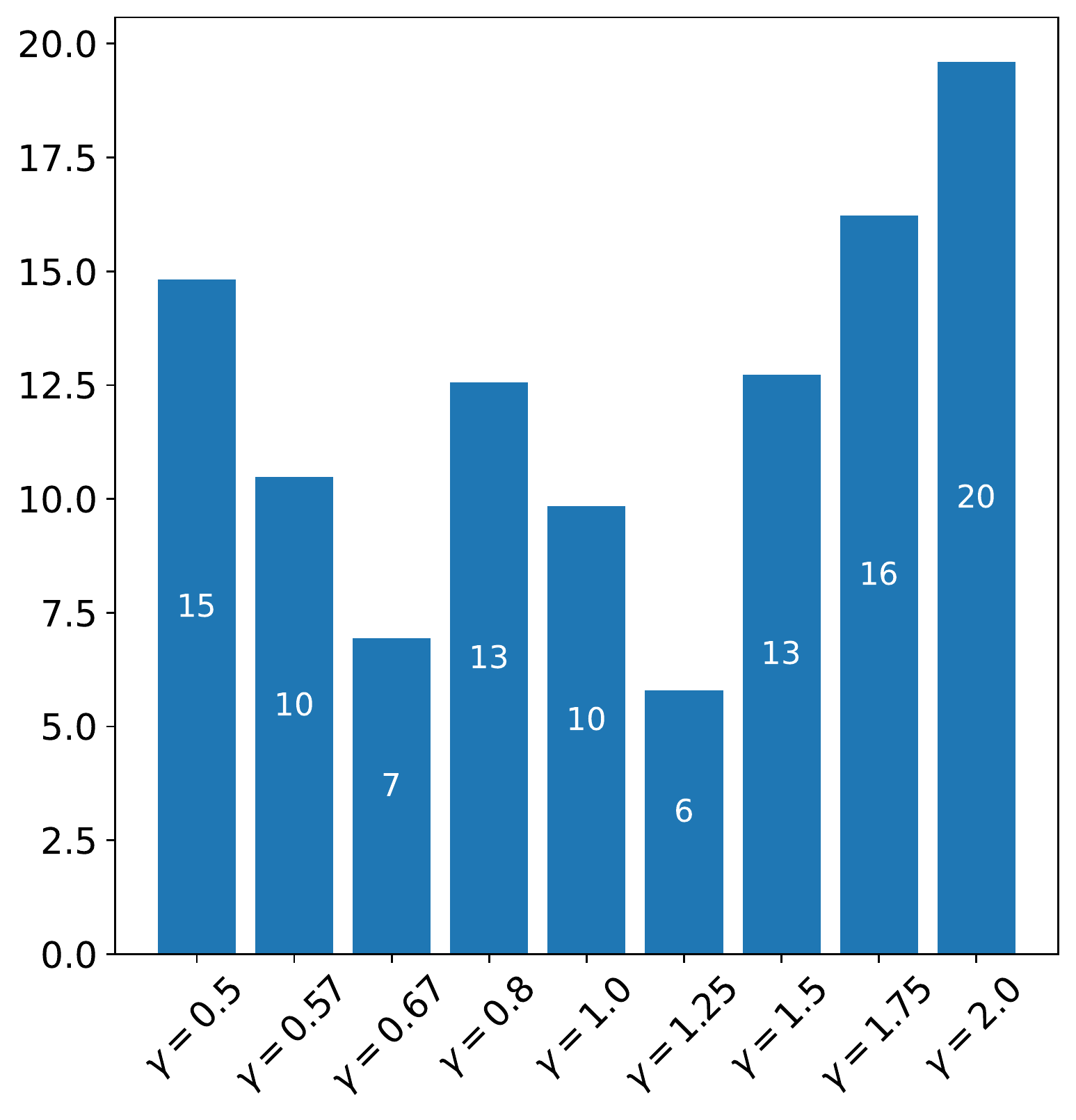}
        \caption{\label{fig:bplot:adtw_ac}Most accurate $\ADTW^{\gamma}$ for $\gamma\in c$}
    \end{subfigure}
    \vspace*{-10pt}
    \caption{\label{fig:bplot_a}
        Counts of the numbers of datasets for which each value of $\gamma$ results in the highest accuracy on the test data.
    }
\end{figure}

\newtext{Figure~\ref{fig:bplot_a} shows the number of datasets for which each exponent results in the  highest accuracy on the test data for each of our NN classifiers and each of the three sets of exponents. It is clear that there is great diversity across datasets in terms of which $\gamma$ is most effective. For $\DTW$, the extremely small $\gamma=0.2$ is desirable for $12\%$ of datasets and the extremely large $\gamma=5.0$ for $8\%$.  }

\newtext{The optimal exponent differs between $\DTW^{\gamma}$ and $\ADTW^{\gamma}$, due to different interactions between the window parameter $w$ for $\DTW$ and the warping penalty parameter $\omega$ for $\ADTW$. We hypothesize that low values of $\gamma$ can serve as a form of pseudo $\omega$, penalizing longer paths by penalizing large numbers of small difference alignments. $\ADTW$ directly penalizes longer paths through its $\omega$ parameter, reducing the need to deploy $\gamma$ in this role.  If this is correct then $\ADTW$ has greater freedom to deploy $\gamma$ to focus more on low or high amplitude effects in the series, as illustrated in Figure~\ref{fig:headline}.}

\subsection{Comparison against non tuned cost functions}\label{subsec:non tuned}

\begin{figure}
    \captionsetup[subfigure]{aboveskip=-2pt,belowskip=7pt}
    \centering
    \begin{subfigure}[b]{0.49\textwidth}
        \includegraphics[width=\textwidth]{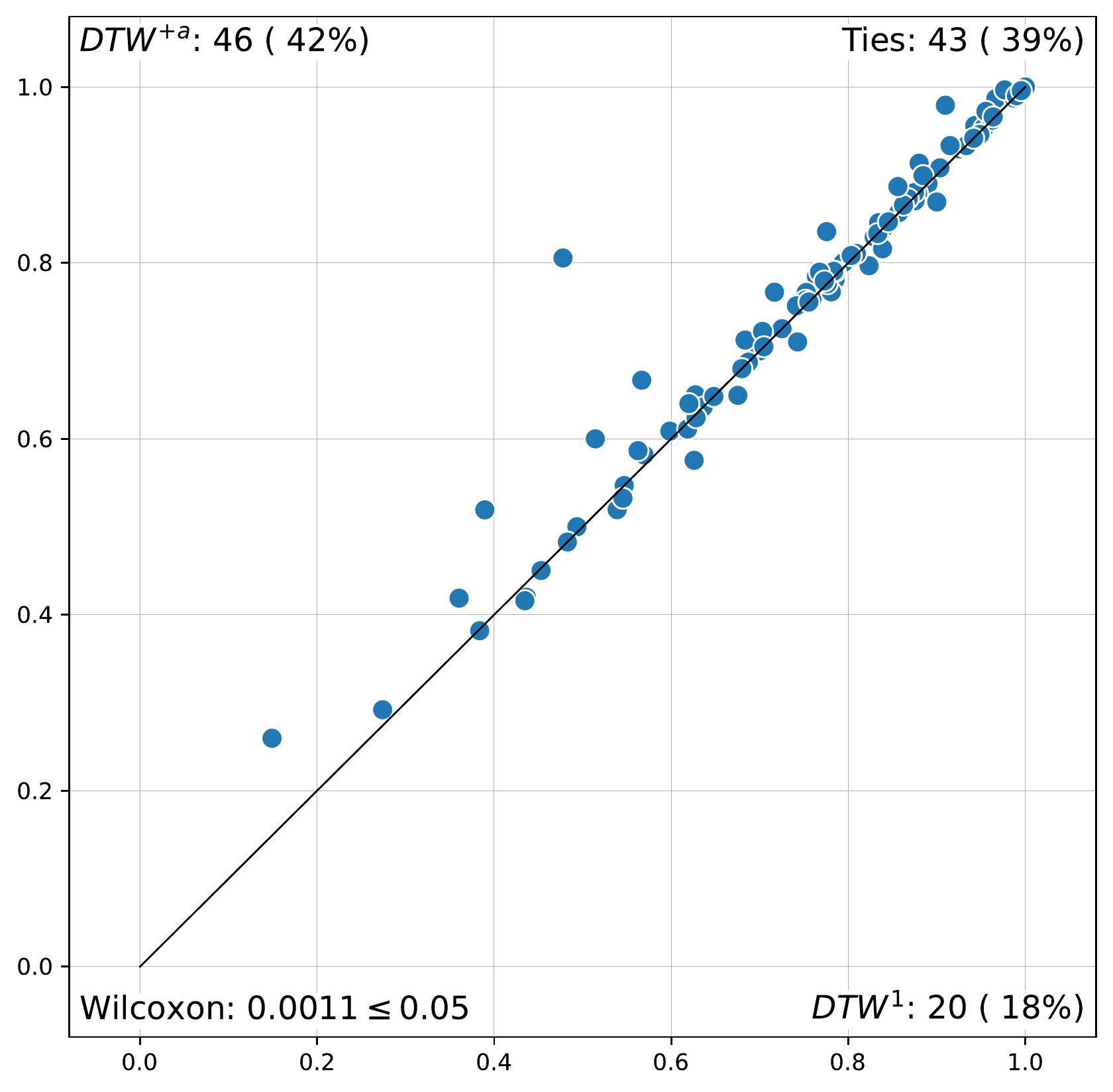}
        \caption{\label{fig:scp:dtw1}$\DTW^{+a}$ vs. $\DTW^{1}$}
    \end{subfigure}
    \hfill
    \begin{subfigure}[b]{0.49\textwidth}
        \includegraphics[width=\textwidth]{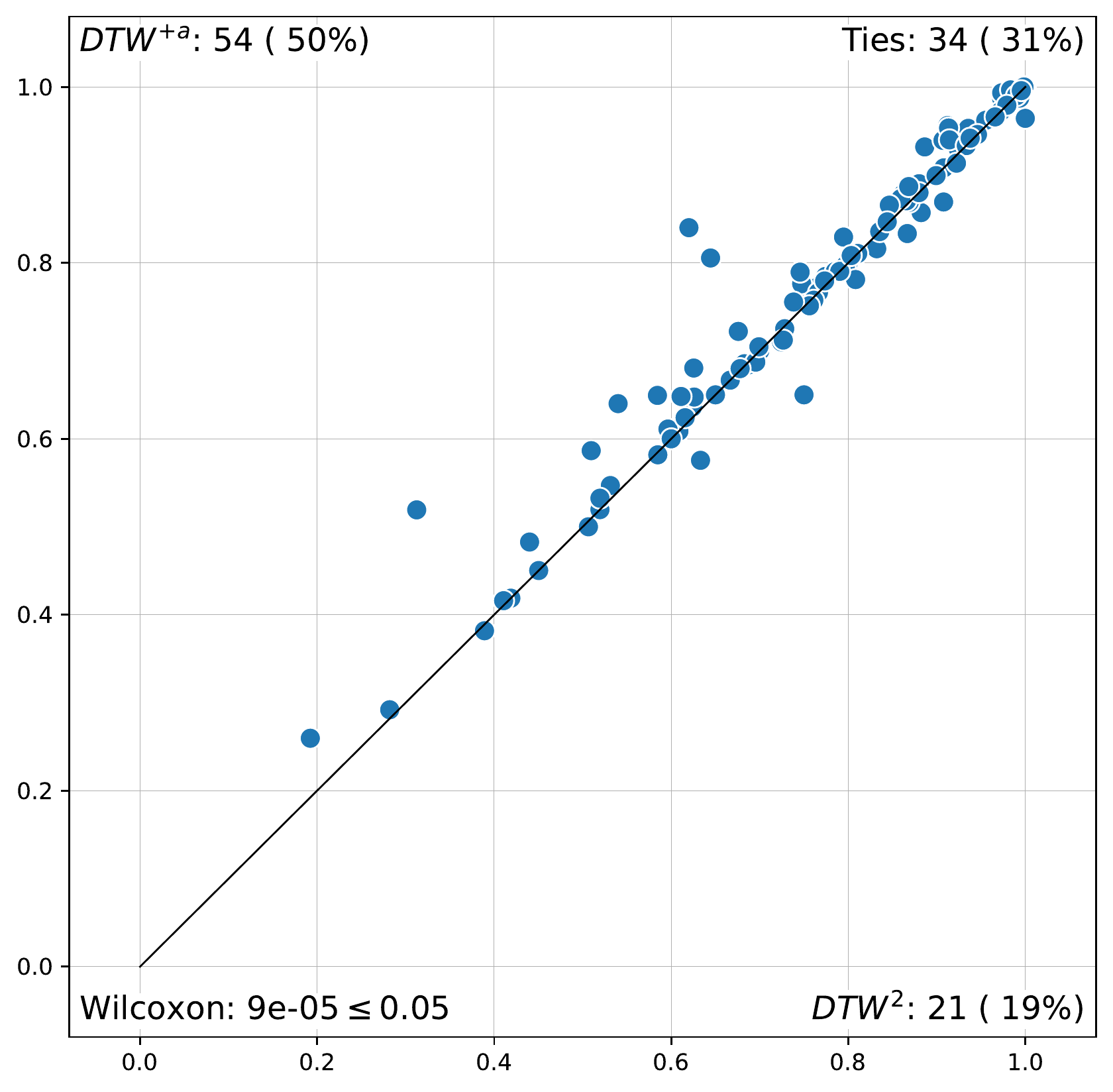}
        \caption{\label{fig:scp:dtw2}$\DTW^{+a}$ vs. $\DTW^{2}$}
    \end{subfigure}
    \vspace*{-10pt}
    \caption{\label{fig:dtw}
                Accuracy scatter plot over the UCR archive comparing
                $\DTW^{+a}$ against $\DTW^{1}$ and $\DTW^{2}$.
                }
\end{figure}

\begin{figure}
    \captionsetup[subfigure]{aboveskip=-2pt,belowskip=7pt}
    \centering
    \begin{subfigure}[b]{0.49\textwidth}
        \includegraphics[width=\textwidth]{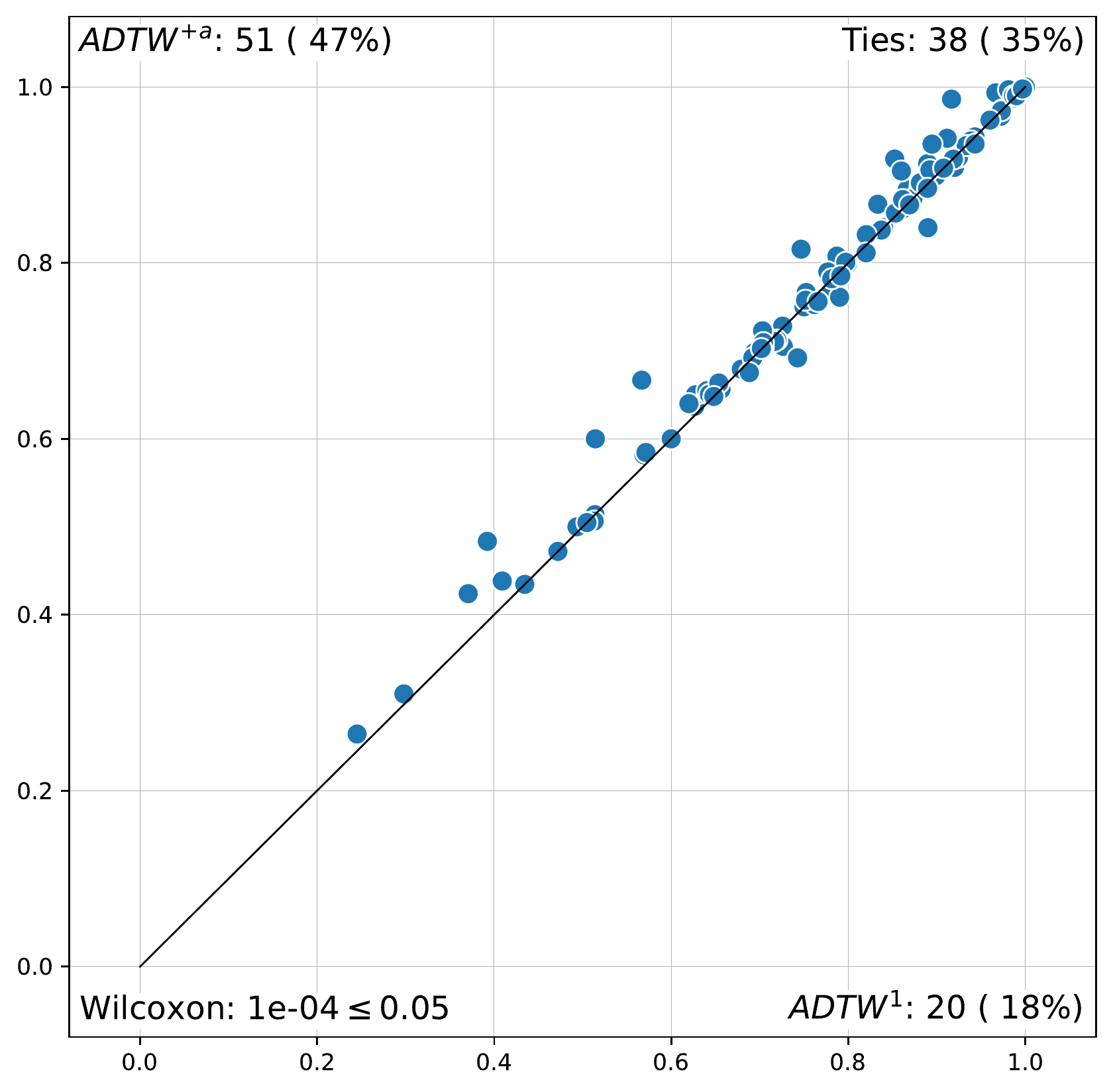}
        \caption{\label{fig:scp:adtw1}$\ADTW^{+a}$ vs. $\ADTW^{1}$}
    \end{subfigure}
    \hfill
    \begin{subfigure}[b]{0.49\textwidth}
        \includegraphics[width=\textwidth]{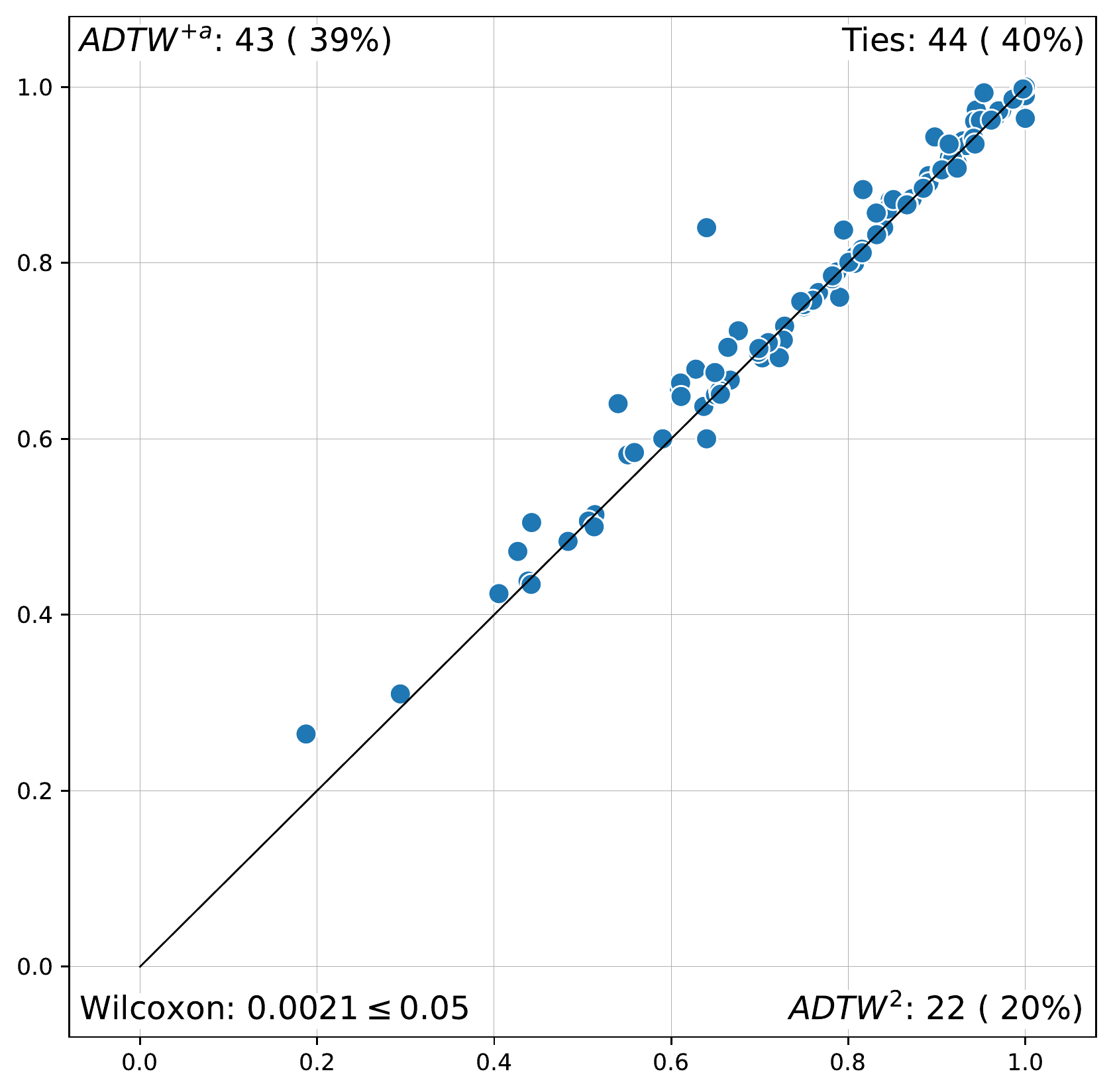}
        \caption{\label{fig:scp:adtw2}$\ADTW^{+a}$ vs. $\ADTW^{2}$}
    \end{subfigure}
    \vspace*{-10pt}
    \caption{\label{fig:adtw}
                Accuracy scatter plot over the UCR archive comparing
                $\ADTW^{+a}$ against $\ADTW^{1}$ and $\ADTW^{2}$.
                }
\end{figure}

\begin{figure}
    \captionsetup[subfigure]{aboveskip=-2pt,belowskip=7pt}
    \centering
    \begin{subfigure}[b]{0.49\textwidth}
        \includegraphics[width=\textwidth]{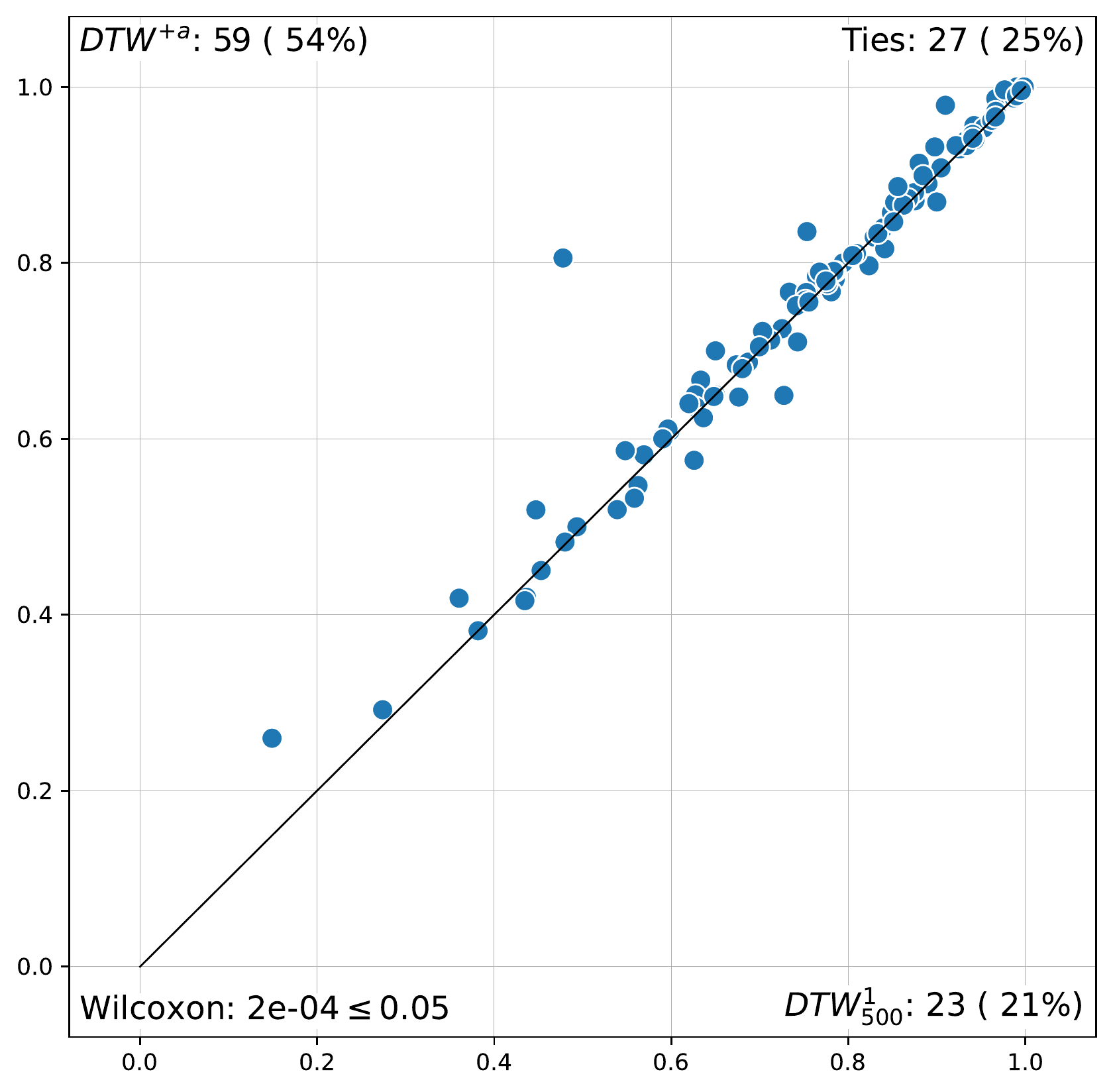}
        \caption{\label{fig:scp:dtw_500}$\DTW^{+a}$ vs $\DTW^{1}_{500}$}
    \end{subfigure}
    \hfill
    \begin{subfigure}[b]{0.49\textwidth}
        \includegraphics[width=\textwidth]{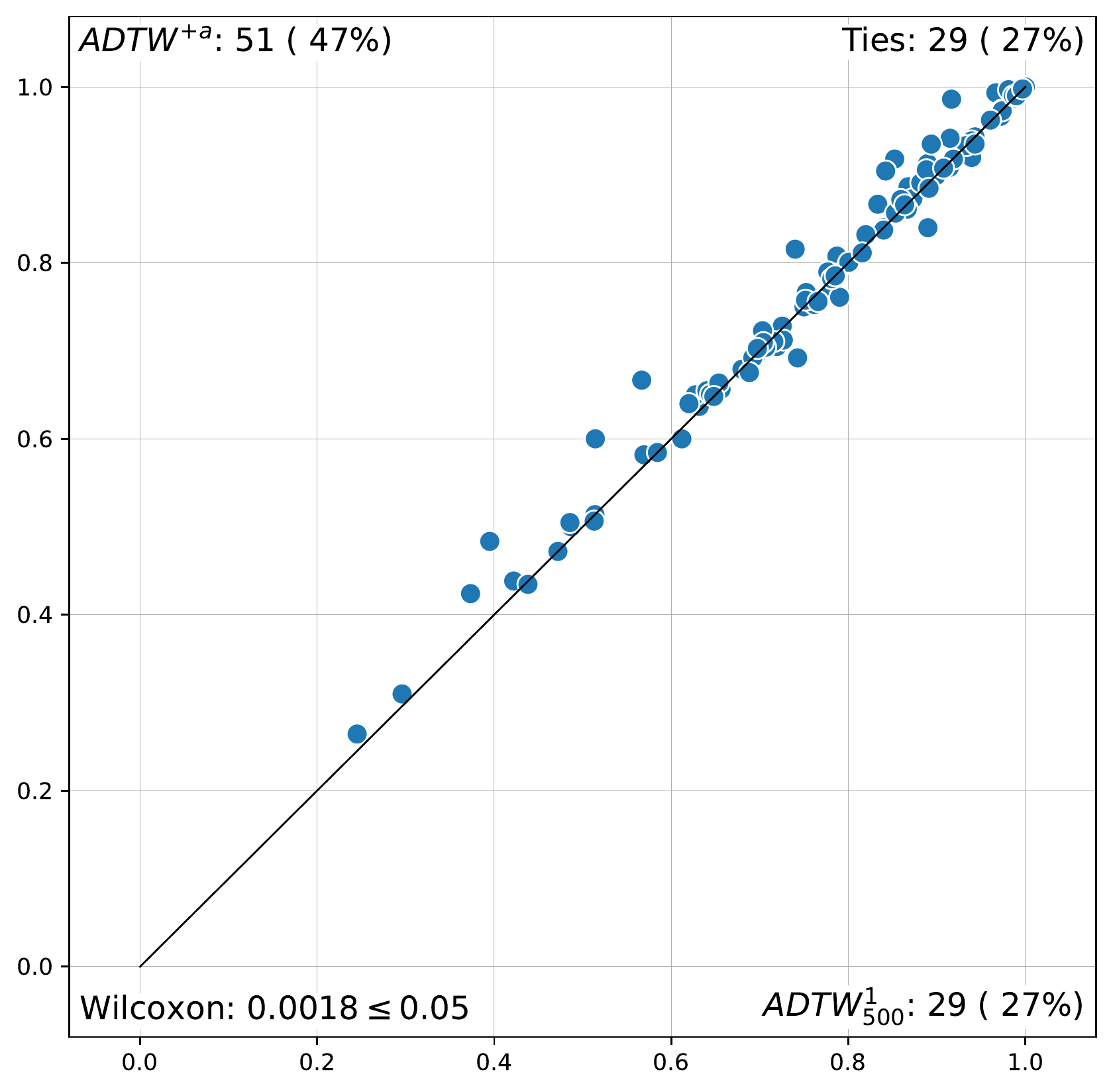}
        \caption{\label{fig:scp:adtw_500}$\ADTW^{+a}$ vs $\ADTW^{1}_{500}$}
    \end{subfigure}
    \vspace*{-10pt}
    \caption{\label{fig:scp:loocv_500}
        Comparison of $\DTW^{+a}$ and $\ADTW^{+a}$ trained over 500 different values
        (5 values for gamma, 100 values for $w$ and $\omega$ per gamma),
        against $\DTW^{1}_{500}$ and $\ADTW^{1}_{500}$ with 500 values for $w$ and $\omega$.
    }
\end{figure}

Figures~\ref{fig:dtw} and~\ref{fig:adtw} present accuracy scatter plots over the UCR archive.
A~dot represents the test accuracy of two classifiers on a dataset.
A dot on the diagonal indicates equal performance for the dataset.
A dot off the diagonal means that the classifier on the corresponding side
(indicated in top left and bottom right corners)
is more accurate than its competitor on this dataset.

On each scatter plot, we also indicate the number of times a classifier is strictly more accurate than its competitor,
the number of ties, and the result of a Wilcoxon signed ranks test indicating
whether the accuracy of the classifiers can be considered significantly different.
Following common practice, we use a significance level of 0.05.

Figures~\ref{fig:dtw} and \ref{fig:adtw}
show that tuning the cost function is beneficial for both $\DTW$ and $\ADTW$,
when compared to both the original cost function $\lambda_1$, and the popular $\lambda_2$.
The Wilcoxon signed ranks test for $\DTW^+$ show that $\DTW^+$ significantly outperforms both $\DTW^1$ and $\DTW^2$.
Similarly, $\ADTW^+$ significantly outperforms both $\ADTW^1$ and $\ADTW^2$.

\subsection{Investigation of the number of parameter values}\label{subsec:parameter numbers}
$\DTW^+$ and $\ADTW^+$ are tuned on 500 parameter options.
To assess whether their improved accuracy is due to an increased number of parameter options
rather than due to the addition of cost tuning per se,
we also compared them against $\DTW^1$ and $\ADTW^1$ also tuned with 500 options
for their parameters $w$ and $\omega$ instead of the usual 100.
Figure~\ref{fig:scp:loocv_500} shows that increasing the number of parameter values available to $\DTW^1$
and $\ADTW^1$ does not alter the advantage of cost tuning.
Note that the warping window $w$ of $\DTW$ is a natural number
for which the range of values that can result in different outcomes is $0\leq{}w\leq{}\tslen-2$.
In consequence, we cannot train $\DTW$ on more than $\tslen-1$ meaningfully different parameter values.
This means that for short series ($\tslen<100$),
increasing the number of possible windows from 100 to 500 has no effect.
$\ADTW$ suffers less from this issue due to the penalty $\omega$ being sampled in a continuous space.
Still, increasing the number of parameter values yields ever diminishing returns,
while increasing the risk of overfitting.
This also means that for a fixed budget of parameter values to be explored,
tuning the cost function as well as $w$ or $\omega$
allows the budget to be \emph{spent} exploring a broader range of possibilities.

\subsection{Comparison against larger tuning sets}

Our experiments so far allow to achieve our primary goal:
to demonstrate that tuning the cost function is beneficial.
We did so with the set of exponents~$a$.
This set is not completely arbitrary
($1$ and $2$ come from current practice, we added their mean $1.5$ and the reciprocals).
However, it remains an open question whether or not it is a reasonable default choice.
Ideally, practitioners need to use expert knowledge to offer
the best possible set of cost functions to choose from for a given application.
In particular, using an alternative form of cost function to $\lambda_\gamma$ could be effective,
although we do not investigate this possibility in this paper.

\begin{figure}
    \captionsetup[subfigure]{aboveskip=-2pt,belowskip=7pt}
    \centering
    \begin{subfigure}[b]{0.49\textwidth}
        \includegraphics[width=\textwidth]{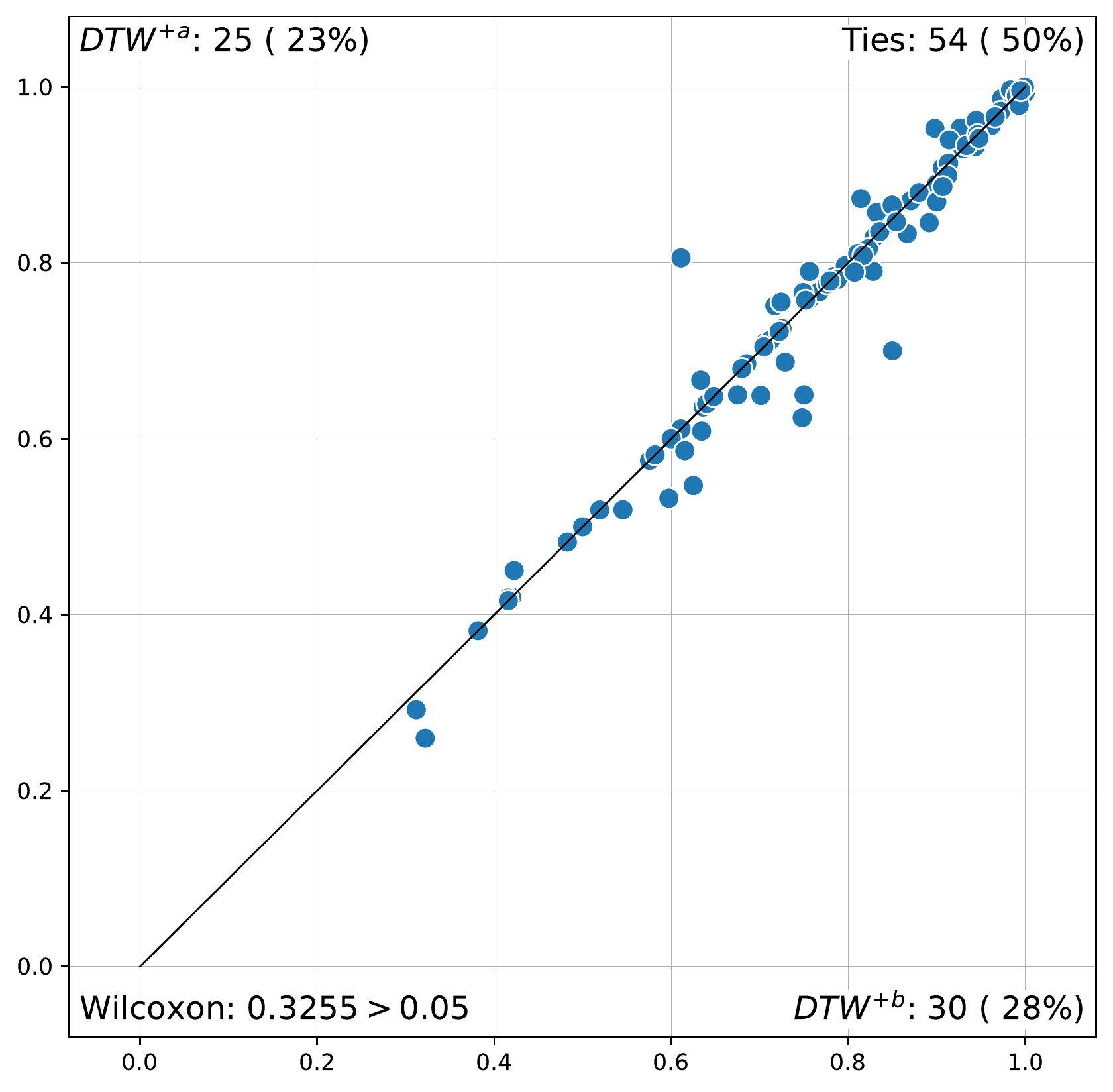}
        \caption{\label{fig:scp:dtw_ab}$\DTW^{+a}$ vs. $\DTW^{+b}$}
    \end{subfigure}
    \hfill
    \begin{subfigure}[b]{0.49\textwidth}
        \includegraphics[width=\textwidth]{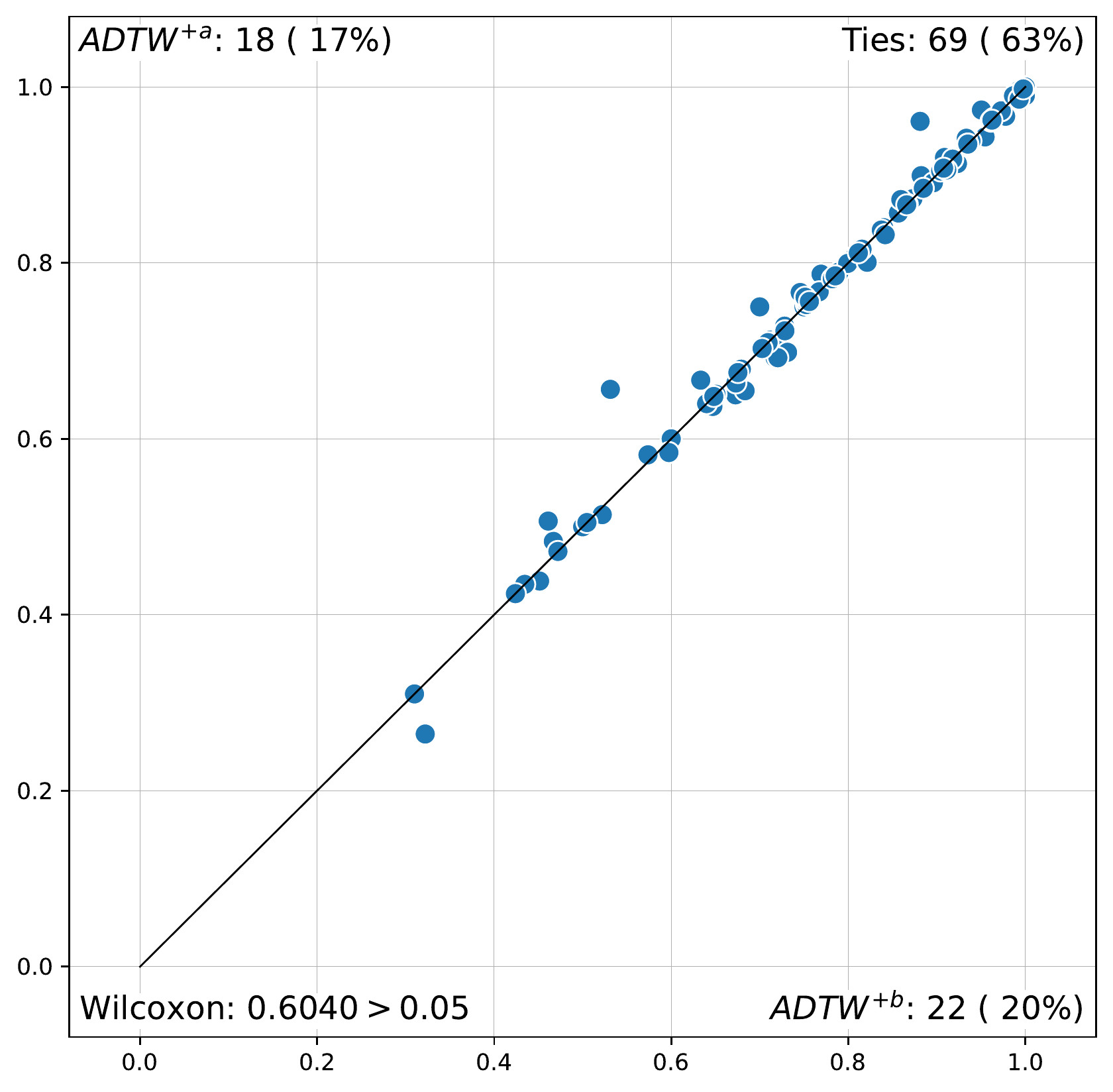}
        \caption{\label{fig:scp:adtw_ab}$\ADTW^{+a}$ vs. $\ADTW^{+b}$}
    \end{subfigure}
    \vspace*{-10pt}
    \caption{\label{fig:scp:setb}Comparison of default exponent set $a$ and \emph{larger} set $b$.
    }
\end{figure}
Figure~\ref{fig:scp:setb} shows the results obtained when using the larger set $b$,
made of 11 values extending $a$ with $3$, $4$, $5$ and their reciprocals.
Compared to $a$, the change benefits $\DTW^+$ (albeit not significantly according to the Wilcoxon test),
at the cost of more than doubling the number of assessed parameter values.
On the other hand, $\ADTW^+$ is mostly unaffected by the change.

\begin{figure}
    \captionsetup[subfigure]{aboveskip=-2pt,belowskip=7pt}
    \centering
    \begin{subfigure}[b]{0.49\textwidth}
        \includegraphics[width=\textwidth]{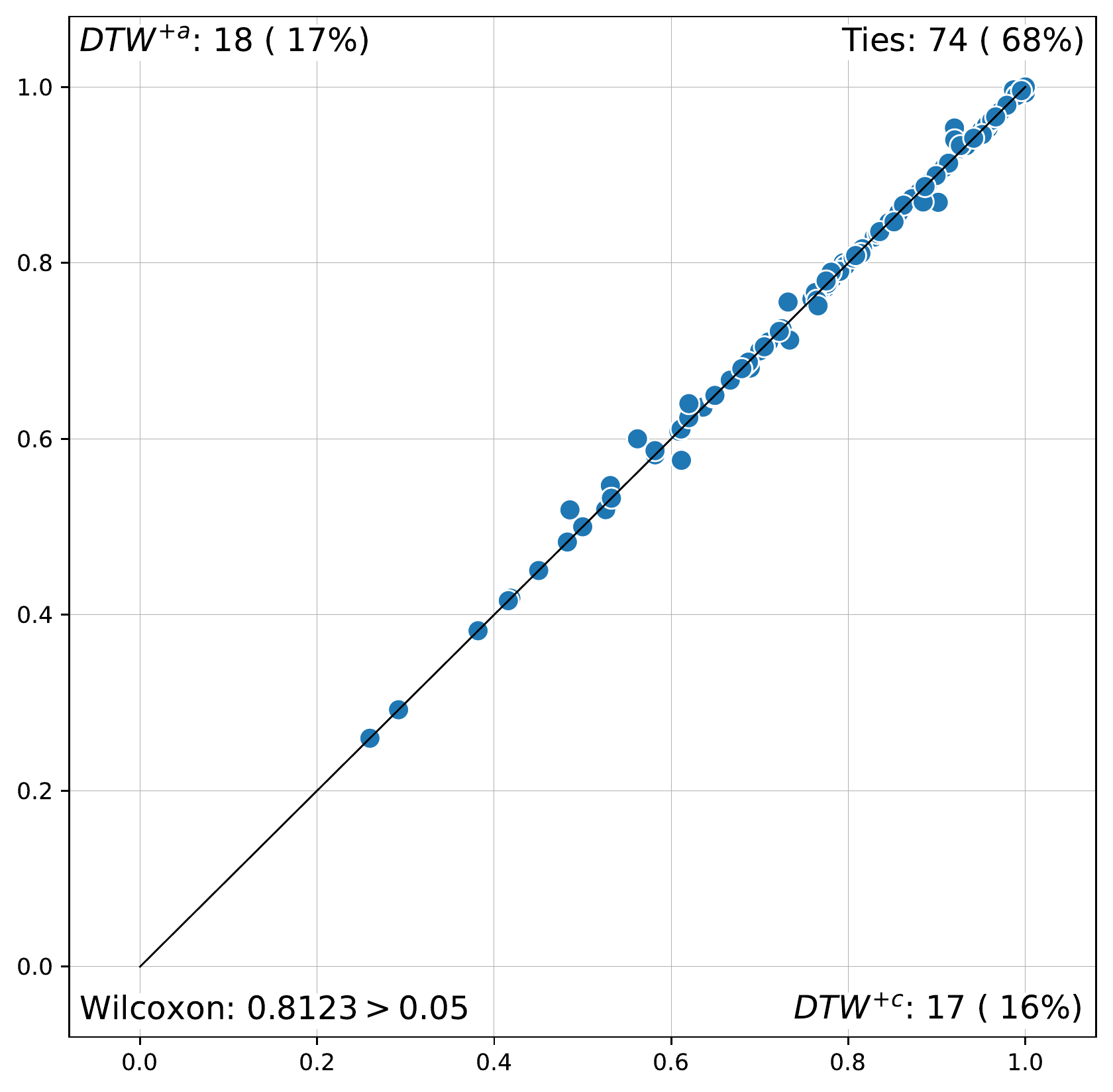}
        \caption{\label{fig:scp:dtw_ac}$\DTW^{+a}$ vs. $\DTW^{+c}$}
    \end{subfigure}
    \hfill
    \begin{subfigure}[b]{0.49\textwidth}
        \includegraphics[width=\textwidth]{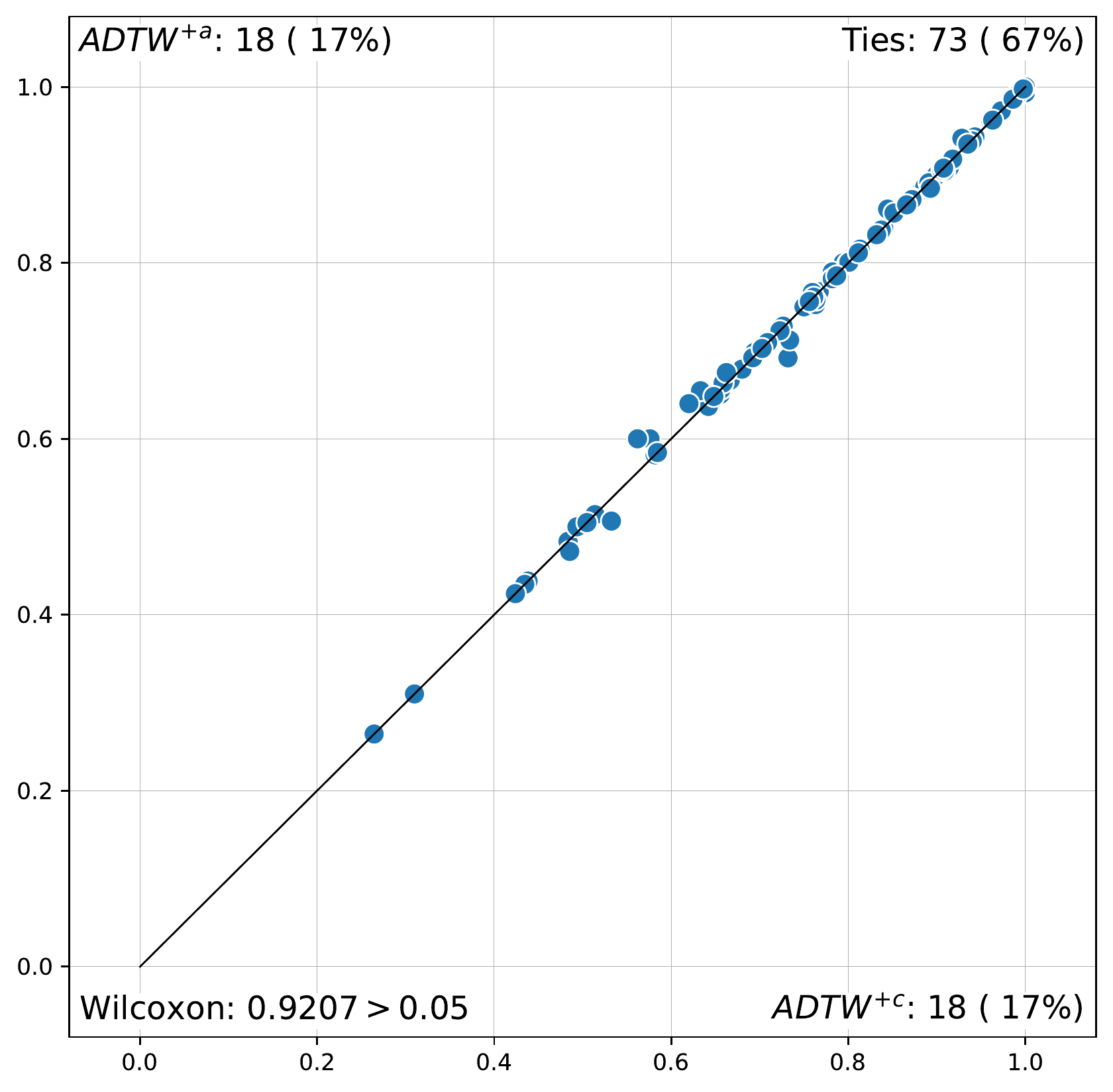}
        \caption{\label{fig:scp:adtw_ac}$\ADTW^{+a}$ vs. $\ADTW^{+c}$}
    \end{subfigure}
    \vspace*{-10pt}
    \caption{\label{fig:scp:setc}Comparison of default exponent set $a$ and \emph{denser} set $c$.
    }
\end{figure}
Figure~\ref{fig:scp:setc} shows the results obtained when using the denser set $c$,
made of 9 values between $0.5$ and $2$.
In this case, neither distance benefits from the change.

\subsection{Runtime}



\newtext{
There is usually a tradeoff between runtime and accuracy for a practical machine learning algorithm. 
Sections \ref{subsec:non tuned} and \ref{subsec:parameter numbers} show that tuning the cost function significantly improves the accuracy of both $\ADTW$ and $\DTW$ in nearest neighbor classification tasks.  
However, this comes at the cost of having more parameters (500 instead of 100 with a single exponent).
TSC using the nearest neighbor algorithm paired with $O(L^2)$ complexity elastic distances are well-known to be computationally expensive, taking hours to days to train \citep{tan2021ultra}.
Therefore, we discuss in this section, the computational details of tuning the cost function $\gamma$ and assess the tradeoff in accuracy gain.
}

\newtext{
We performed a runtime analysis by recording the total time taken to train and test both $\DTW$ and $\ADTW$ for each $\gamma$ from the default set $a$.
Our experiments were coded in C++ and parallelised on a machine with 32 cores and AMD EPYC-Rome 2.2Ghz Processor for speed.
The C++ \texttt{pow} function that supports exponentiation of arbitrary values is computationally demanding.  Hence, we use specialized code to calculate the exponents $0.5$, $1.0$ and $2.0$ efficiently, using \texttt{sqrt} for $0.5$, \texttt{abs} for $1.0$ and multiplication for $2.0$.
}

\newtext{
Figure \ref{fig:timing_train}
\begin{figure}
    \centering
    \begin{subfigure}[b]{0.49\textwidth}
        \includegraphics[width=\textwidth]{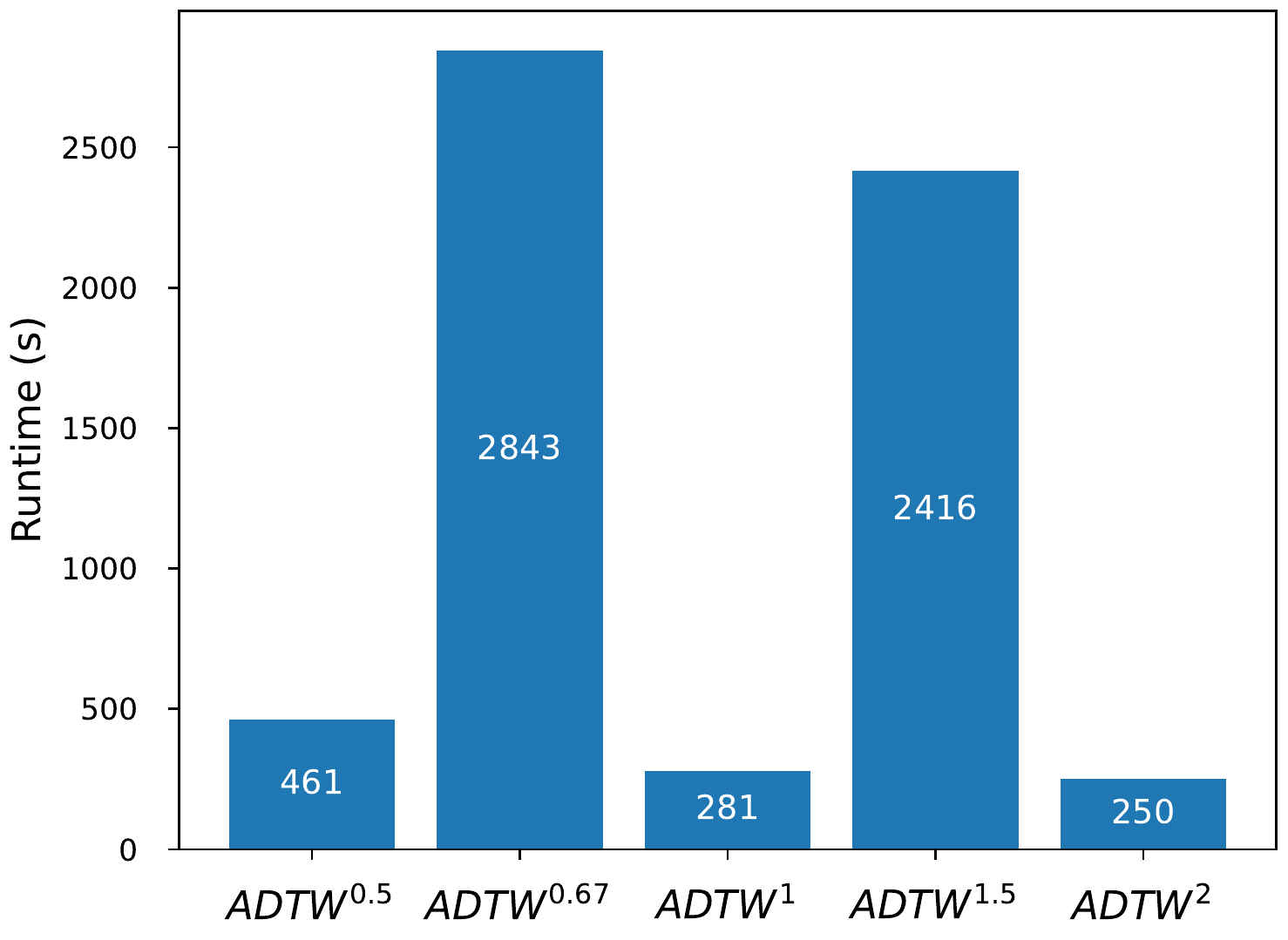}
        \caption{\label{fig:scp:timing_ADTW_train}$\ADTW$}
    \end{subfigure}
    \hfill
    \begin{subfigure}[b]{0.49\textwidth}
        \includegraphics[width=\textwidth]{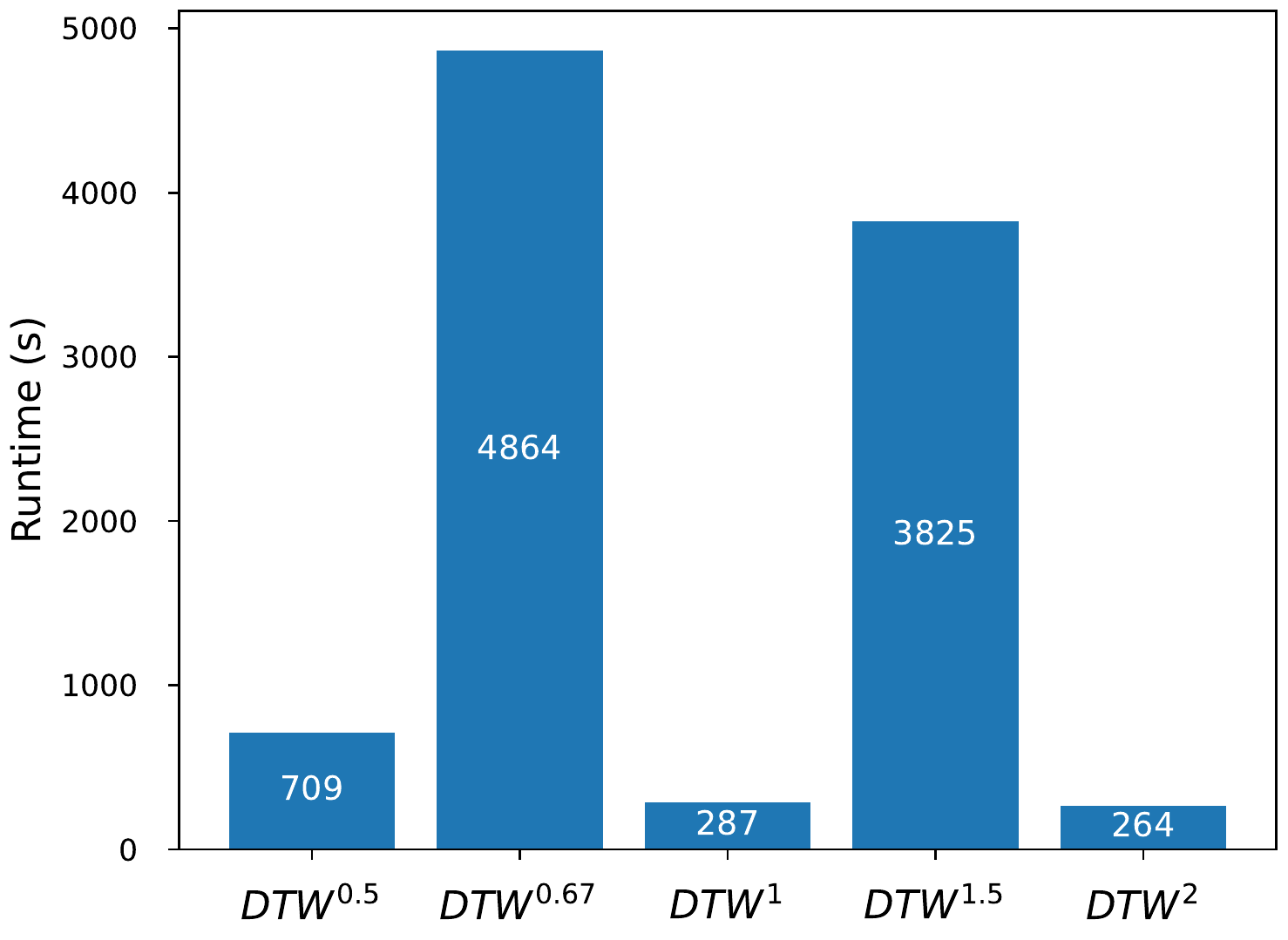}
        \caption{\label{fig:scp:timing_DTW_test}$\DTW$}
    \end{subfigure}
    \caption{\label{fig:timing_train}
                LOOCV train time in seconds on the UCR Archive (109 datasets) of each distance, per exponent. These timings are done on a machine with 32 cores and AMD EPYC-Rome 2.2Ghz Processor.
                }
\end{figure}%
shows the LOOCV training time for both $\ADTW$ and $\DTW$ on each $\gamma$, while Figure \ref{fig:timing_test}
\begin{figure}
    \centering
    \begin{subfigure}[b]{0.49\textwidth}
        \includegraphics[width=\textwidth]{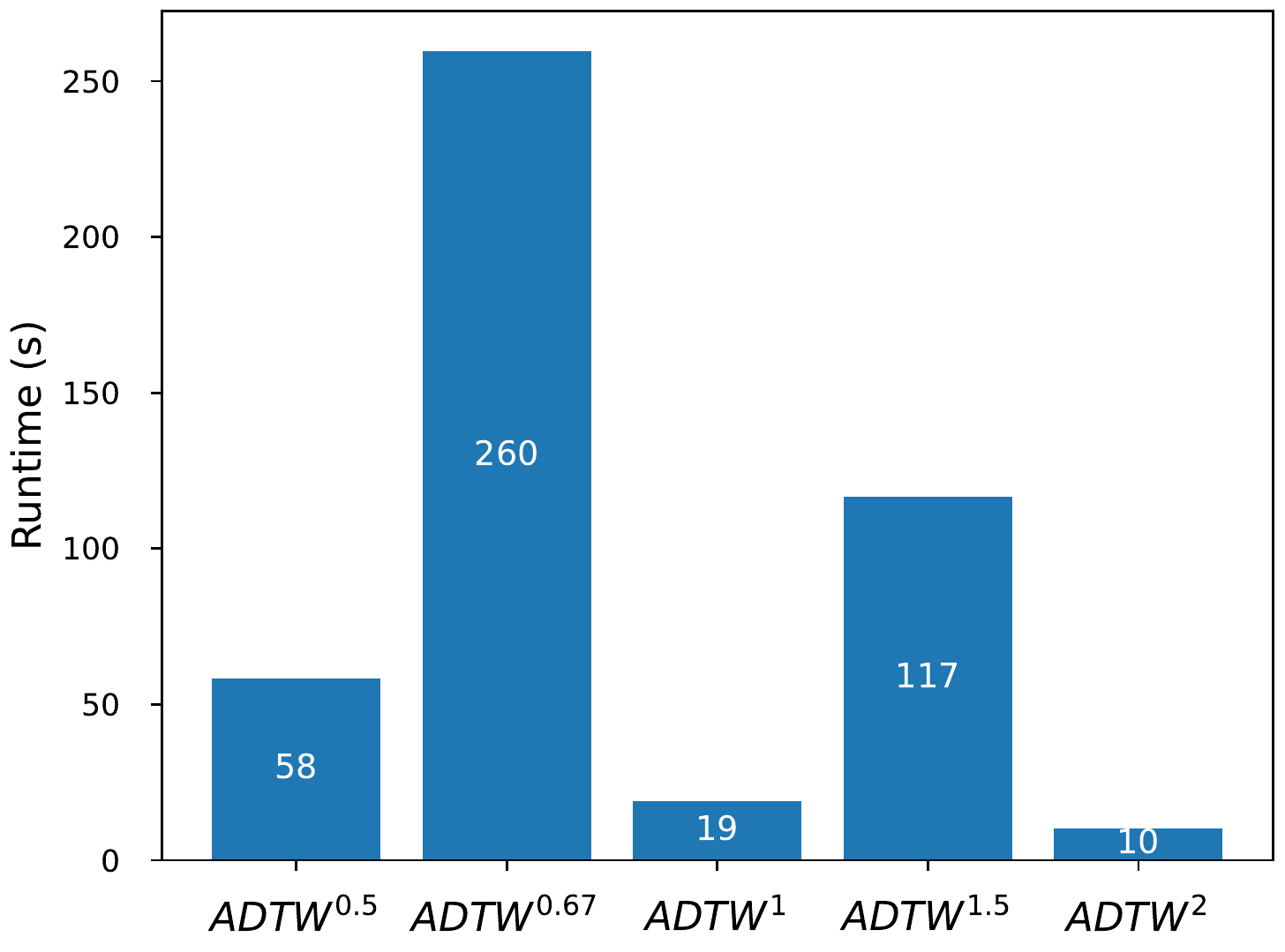}
        \caption{\label{fig:scp:timing_ADTW_test}$\ADTW$}
    \end{subfigure}
    \hfill
    \begin{subfigure}[b]{0.49\textwidth}
        \includegraphics[width=\textwidth]{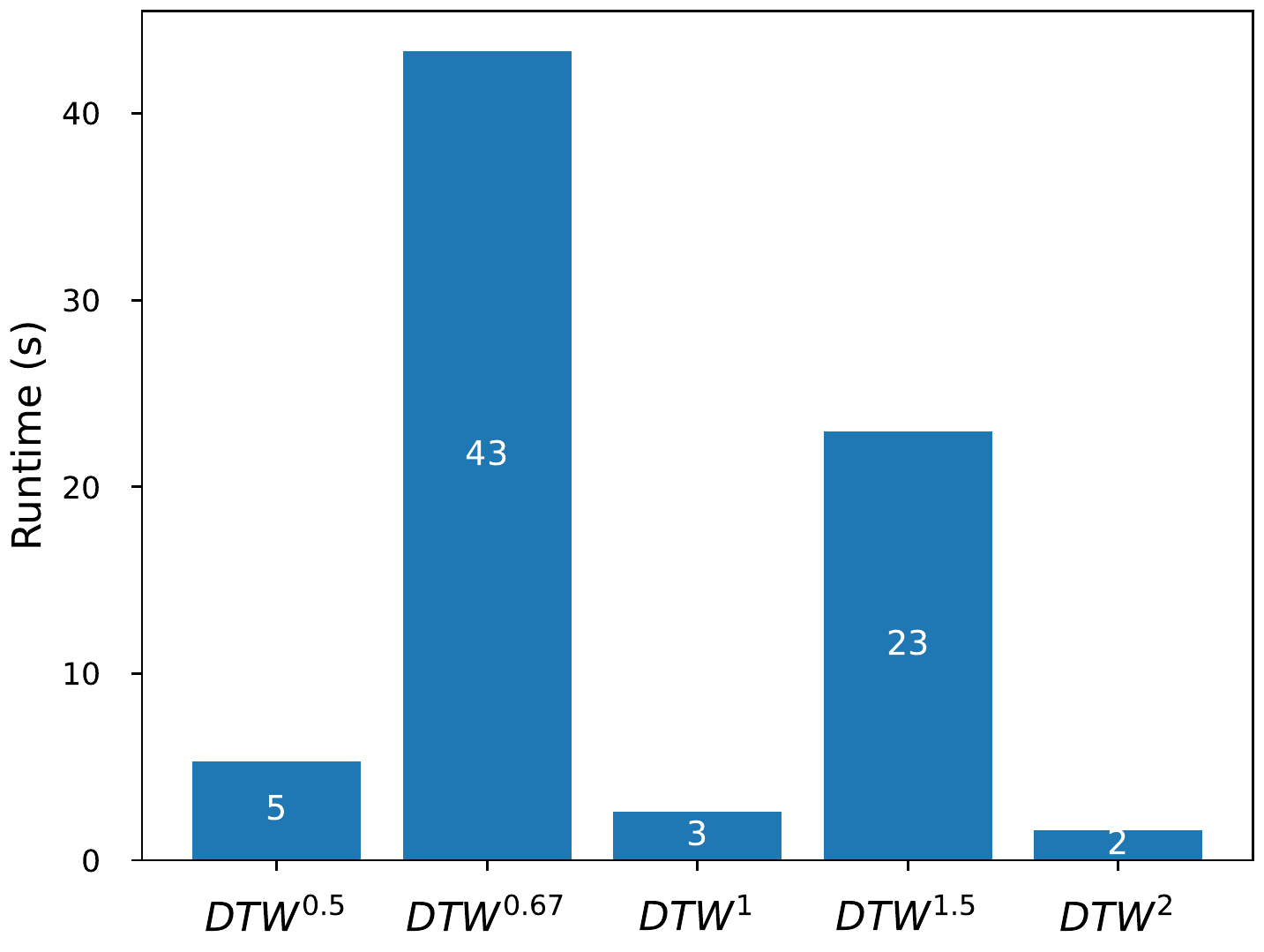}
        \caption{\label{fig:scp:timing_DTW_test}$\DTW$}
    \end{subfigure}
    \caption{\label{fig:timing_test}
                Test time in seconds on the UCR Archive (109 datasets) of each distance, per exponent.
                These timings are done on a machine with 32 cores and AMD EPYC-Rome 2.2Ghz Processor.
                }
\end{figure}%
shows the test time. 
The runtimes for $\gamma{=}0.67$ and $\gamma{=}1.5$ are both substantially longer than those of the specialized exponents.
The~total time to tune the cost function and their parameters on 109 UCR time series datasets are 6250.94 (2 hours) and 9948.98 (3 hours) seconds for $\ADTW$ and $\DTW$ respectively.
This translates to $\ADTW^+$ and $\DTW^+$ being approximately 25 and 38 times slower than the baseline setting with $\gamma{=}2$. 
Potential strategies for reducing these substantial computational burdens are to only use exponents that admit efficient computation, such as powers of 2 and their reciprocals. Also, the parameter tuning for $w$ and $\omega$ in these experiment does not exploit the substantial speedups of recent $\DTW$ parameter search methods \citep{tan2021ultra}.
Despite being slower than both distances at $\gamma{=}2$, completing the training of all 109 datasets under 3 hours is still significantly faster than many other TSC algorithms \citep{tan2022multirocket,middlehurst2021hive}
}

\subsection{Noise}

\newtext{As $\gamma$ alters $\DTW$'s relative responsiveness to different magnitudes of effect in a pair of series, it is credible that tuning it may be helpful when the series are noisy. On~one hand, higher values of $\gamma$ will help focus on large magnitude effects, allowing $\DTW$ to pay less attention to smaller magnitude effects introduced by noise. On~the other hand, lower values of $\gamma$ will increase focus on small magnitude effects introduced by noise, increasing the ability of $\DTW^\gamma$ to penalize long warping paths that align sets of similar values.}

\newtext{To examine these questions we created two variants of each of the UCR datasets.  
For the first dataset we added moderate random noise, adding $0.1 \times \mathcal{N}(0, \sigma)$ to each time step, where $\sigma$ is the standard deviation in the values in the series.
For~the second dataset (substantial noise) we added $\mathcal{N}(0, \sigma)$ to each time step.}

\begin{figure}
    \includegraphics[width=\textwidth,trim={0 0 0 10mm},clip]{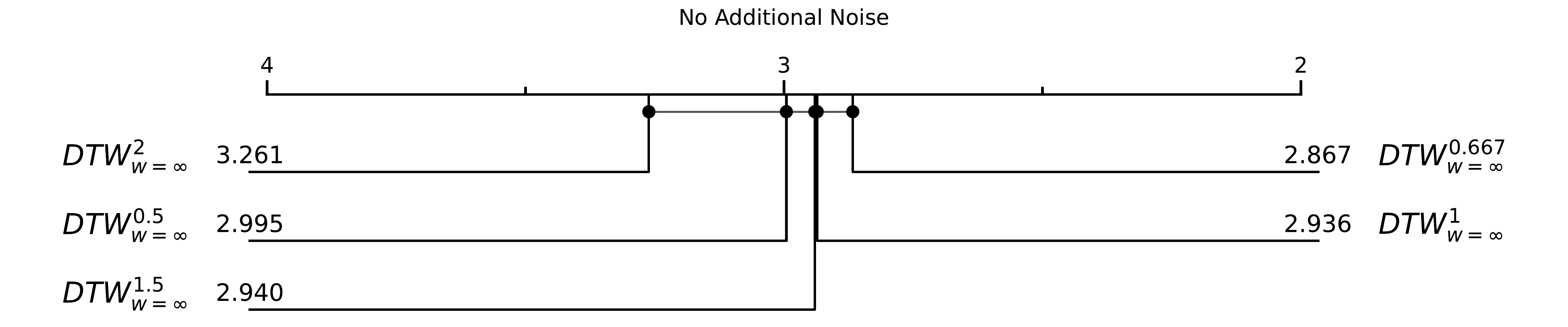}
    \caption{\label{fig:scp:cd_DTWFull_noise_00}
        Critical Difference Diagram for $\DTWi$ on the UCR Archive (109 datasets) with no additional noise.
    }
\end{figure}

\begin{figure}
    \includegraphics[width=\textwidth,trim={0 0 0 10mm},clip]{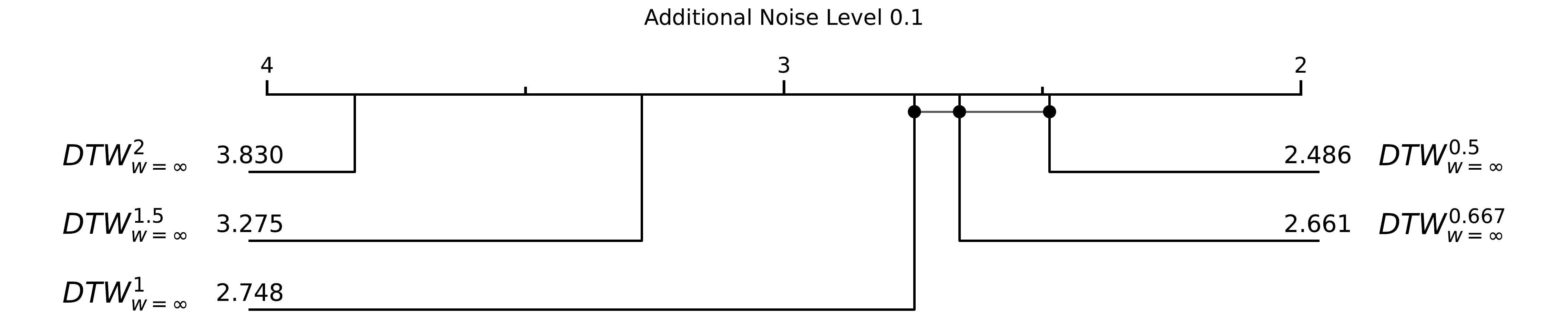}
    \caption{\label{fig:scp:cd_DTWFull_noise_01}
        Critical Difference Diagram for $\DTWi$ on the UCR Archive (109 datasets)
        with moderate additional noise.
    }
\end{figure}

\begin{figure}
    \includegraphics[width=\textwidth,trim={0 0 0 10mm},clip]{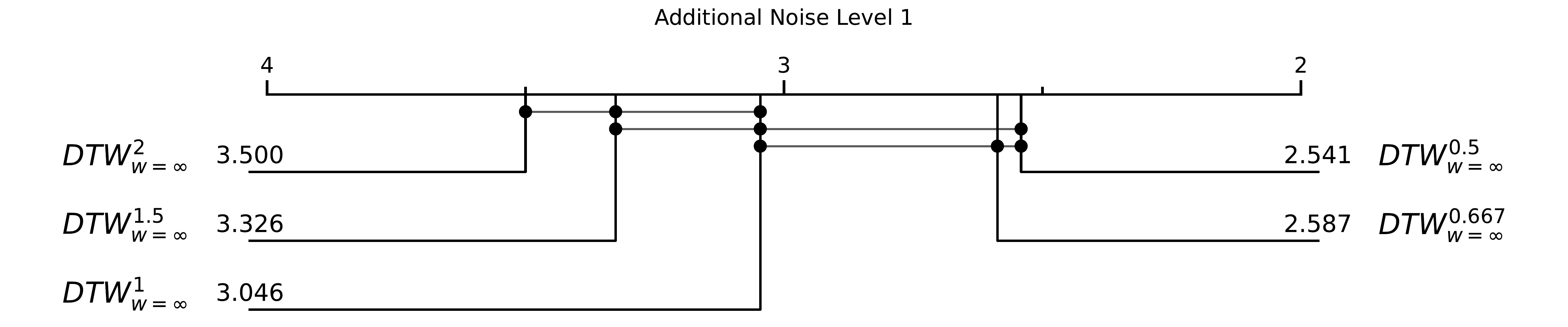}
    \caption{\label{fig:scp:cd_DTWFull_noise_1}
        Critical Difference Diagram for $\DTWi$ on the UCR Archive (109 datasets)
        with substantial additional noise.
    }
\end{figure}

\newtext{The results for $\DTW^\gamma_{w=\infty}$ ($\DTW$ with no window) are presented in Figures~\ref{fig:scp:cd_DTWFull_noise_00} (no additional noise), \ref{fig:scp:cd_DTWFull_noise_01} (moderate additional noise) and \ref{fig:scp:cd_DTWFull_noise_1} (substantial additional noise). Each figure presents a critical difference diagram. $\DTW^\gamma$ has been applied with all 109 datasets at each $\gamma\in a$. For each dataset, the performance for each $\gamma$ is ranked in descending order on accuracy. The diagram presents the mean rank for each $\DTW^\gamma$ across all datasets, with the best mean rank listed rightmost. Lines connect results that are not significantly different at the 0.05 level on a Wilcoxon singed rank test (for each line, the settings indicated with dots are not significantly different). With no additional noise, no setting of $\gamma$ significantly outperforms the others. With a moderate amount of noise, the three lower values of $\gamma$ significantly outperform the higher values.  We hypothesize that this is as a result of $\DTW$ using the small differences introduced by noise to penalize excessively long warping paths. With high noise, the three lowest $\gamma$ are still significantly outperforming the highest level, but the difference in ranks is closing.  We hypothesize that this is due to increasingly large differences in value being the only ones that remain meaningful, and hence increasingly needing to be emphasized.  }

\begin{figure}
    \includegraphics[width=\textwidth,trim={0 0 0 10mm},clip]{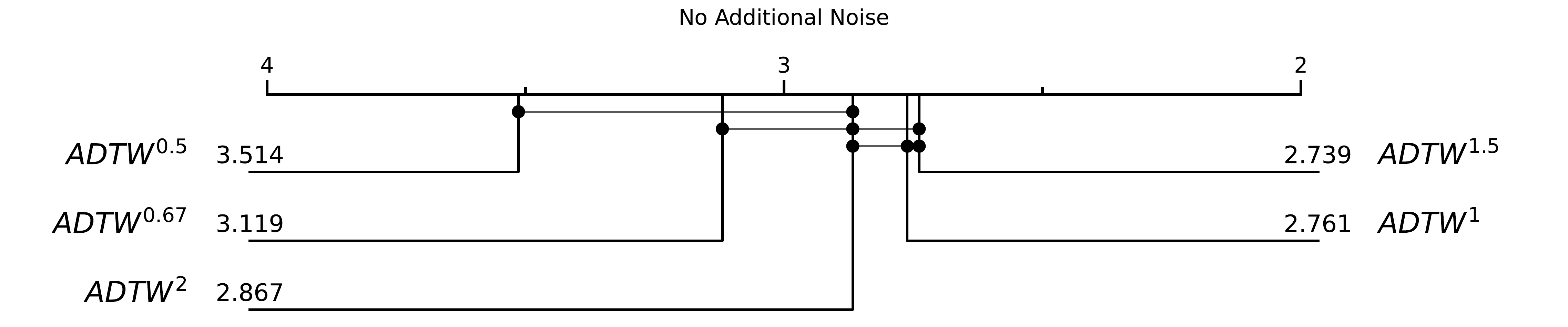}
    \caption{\label{fig:scp:cd_ADTW_noise_00}
        Critical Difference Diagram for $\ADTW$ on the UCR Archive (109 datasets) with no additional noise.
    }
\end{figure}

\begin{figure}
    \includegraphics[width=\textwidth,trim={0 0 0 10mm},clip]{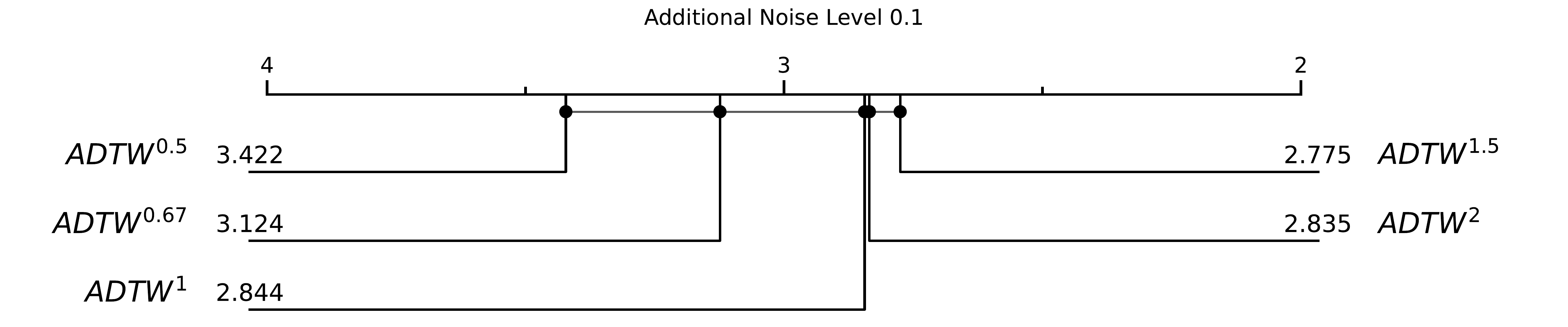}
    \caption{\label{fig:scp:cd_ADTW_noise_01}
        Critical Difference Diagram for $\ADTW$ on the UCR Archive (109 datasets)
        with moderate additional noise.
    }
\end{figure}

\begin{figure}
    \includegraphics[width=\textwidth,trim={0 0 0 10mm},clip]{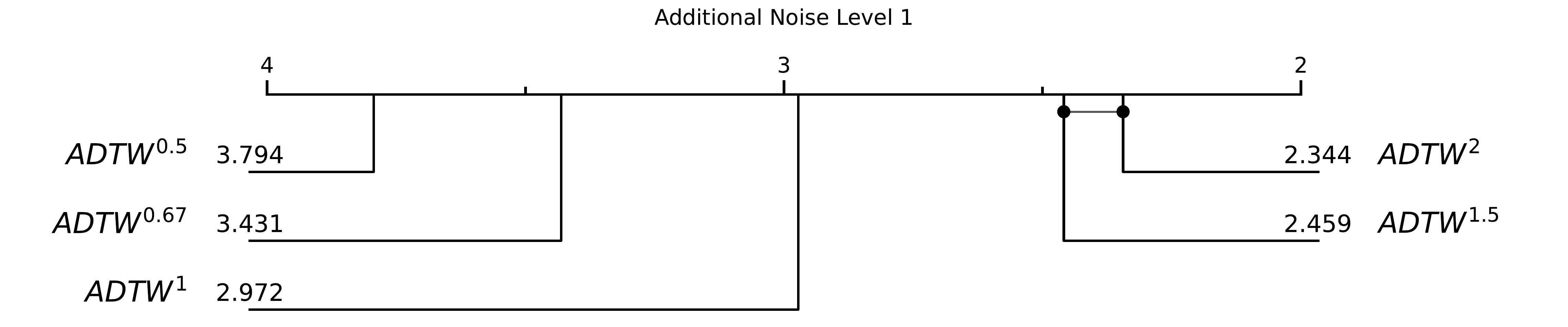}
    \caption{\label{fig:scp:cd_ADTW_noise_1}
        Critical Difference Diagram for $\ADTW$ on the UCR Archive (109 datasets)
        with substantial additional noise.
    }
\end{figure}

\newtext{The results for $\ADTW^\gamma$ are presented in Figures~\ref{fig:scp:cd_ADTW_noise_00} (no additional noise), \ref{fig:scp:cd_ADTW_noise_01} (moderate additional noise) and \ref{fig:scp:cd_ADTW_noise_1} (substantial additional noise). With no additional noise, $\gamma$ values of $1.5$ and $1.0$ both significantly outperform $0.5$. With a moderate amount of noise, $\gamma=2.0$ increases its rank and no value significantly outperforms any other.  With substantial noise, the two highest $\gamma$ significantly outperform all others.  As $\ADTW$ has a direct penalty for longer paths, we hypothesize that this gain in rank for the highest $\gamma$ is due to $\ADTW$ placing higher emphasis on larger differences that are less likely to be the result of noise. }

\begin{figure}
    \includegraphics[width=\textwidth,trim={0 0 0 10mm},clip]{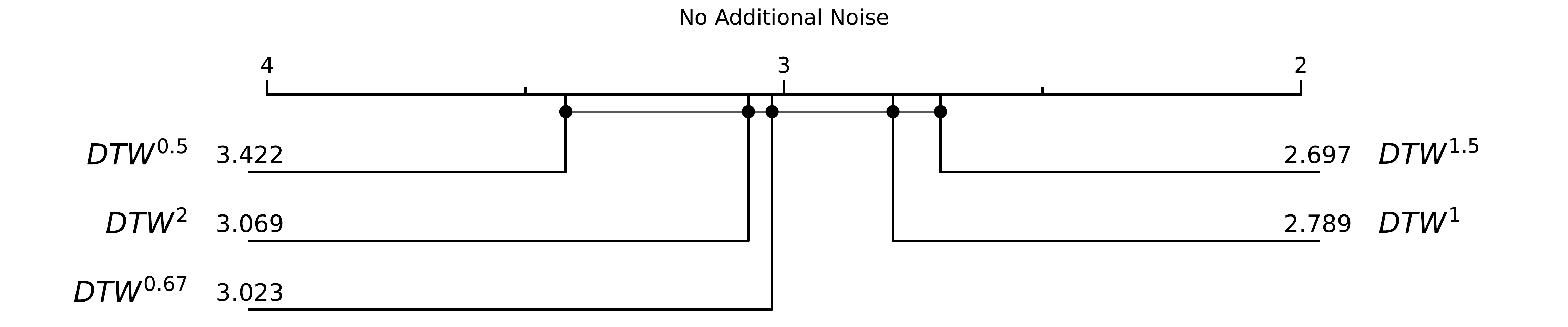}
    \caption{\label{fig:scp:cd_DTW_noise_00}
        Critical Difference Diagram for $\DTW$ on the UCR Archive (109 datasets) with no additional noise.
    }
\end{figure}

\begin{figure}
    \includegraphics[width=\textwidth,trim={0 0 0 10mm},clip]{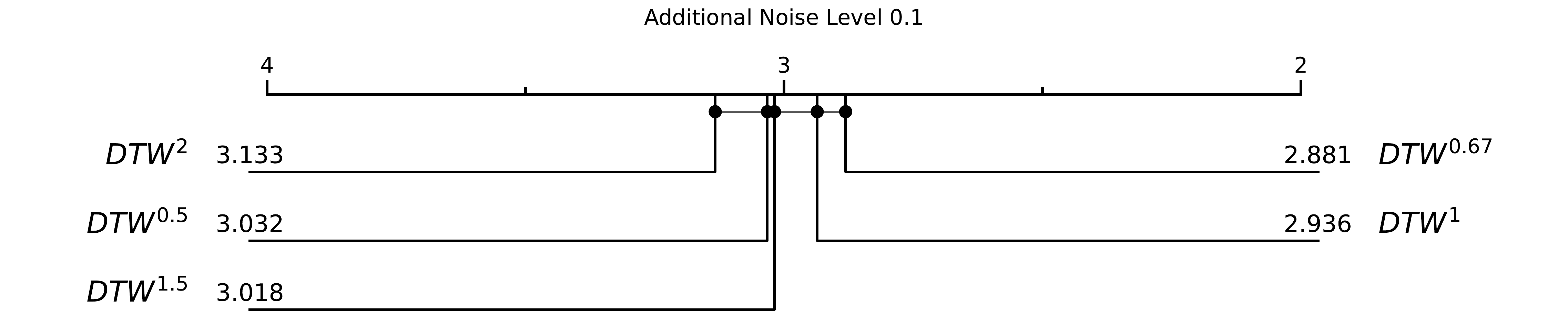}
    \caption{\label{fig:scp:cd_DTW_noise_01}
        Critical Difference Diagram for $\DTW$ on the UCR Archive (109 datasets)
        with lomoderate additional noise.
    }
\end{figure}

\begin{figure}
    \includegraphics[width=\textwidth,trim={0 0 0 10mm},clip]{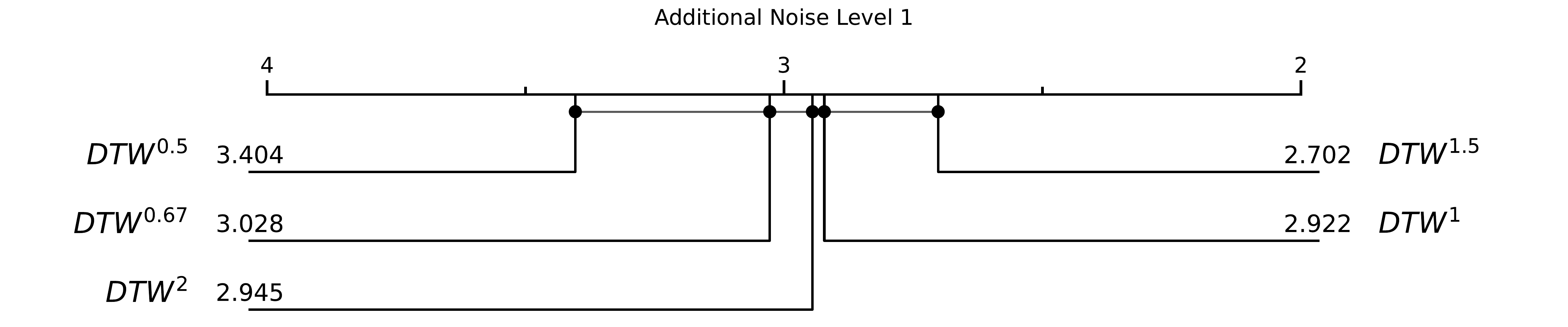}
    \caption{\label{fig:scp:cd_DTW_noise_1}
        Critical Difference Diagram for $\DTW$ on the UCR Archive (109 datasets)
        with substantial additional noise.
    }
\end{figure}

\newtext{The results for $\DTW$ with window tuning are presented in Figures~\ref{fig:scp:cd_DTW_noise_00} (no additional noise), \ref{fig:scp:cd_DTW_noise_01} (moderate additional noise) and \ref{fig:scp:cd_DTW_noise_1} (substantial additional noise).  No setting of $\gamma$ has a significant advantage over any other at any level of noise. We hypothesize that this is because the constraint a window places on how far a warping path can deviate from the diagonal only partially restricts path length, allowing any amount of warping within the window.  Thus, $\DTW$ still benefits from the use of low $\gamma$ to penalize excessive path warping that might otherwise fit noise.  However, it is also subject to a countervailing pressure towards higher values of $\gamma$ in order to focus on larger differences in values that are less likely to be the result of noise.}

\newtext{It is evident from these results that  $\gamma$ interacts in different ways with the $w$ and $\omega$ parameters of $\DTW$ and $\ADTW$ with respect to noise. For $\ADTW$, larger values of $\gamma$ are an effective mechanism to counter noisy series.}

\subsection{Comparing $\DTW^+$ vs $\ADTW^+$}

\begin{figure}
    \captionsetup[subfigure]{aboveskip=-2pt,belowskip=7pt}
    \centering
    \begin{subfigure}[b]{0.49\textwidth}
        \includegraphics[width=\textwidth]{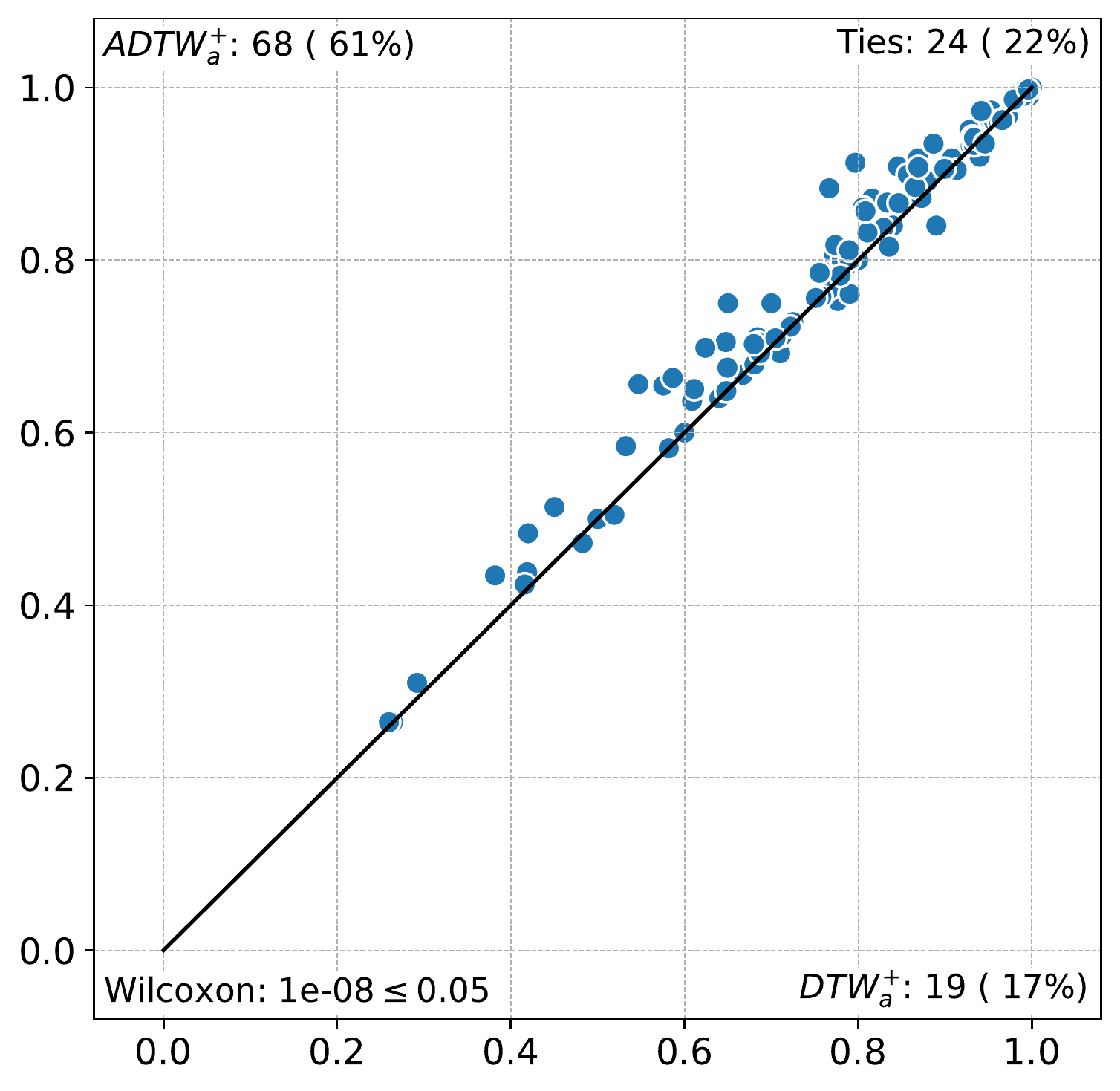}
        \caption{\label{fig:scp:adtw_dtw_a}$\ADTW^{+a}$ vs. $\DTW^{+a}$}
    \end{subfigure}
    \hfill
    \begin{subfigure}[b]{0.49\textwidth}
        \includegraphics[width=\textwidth]{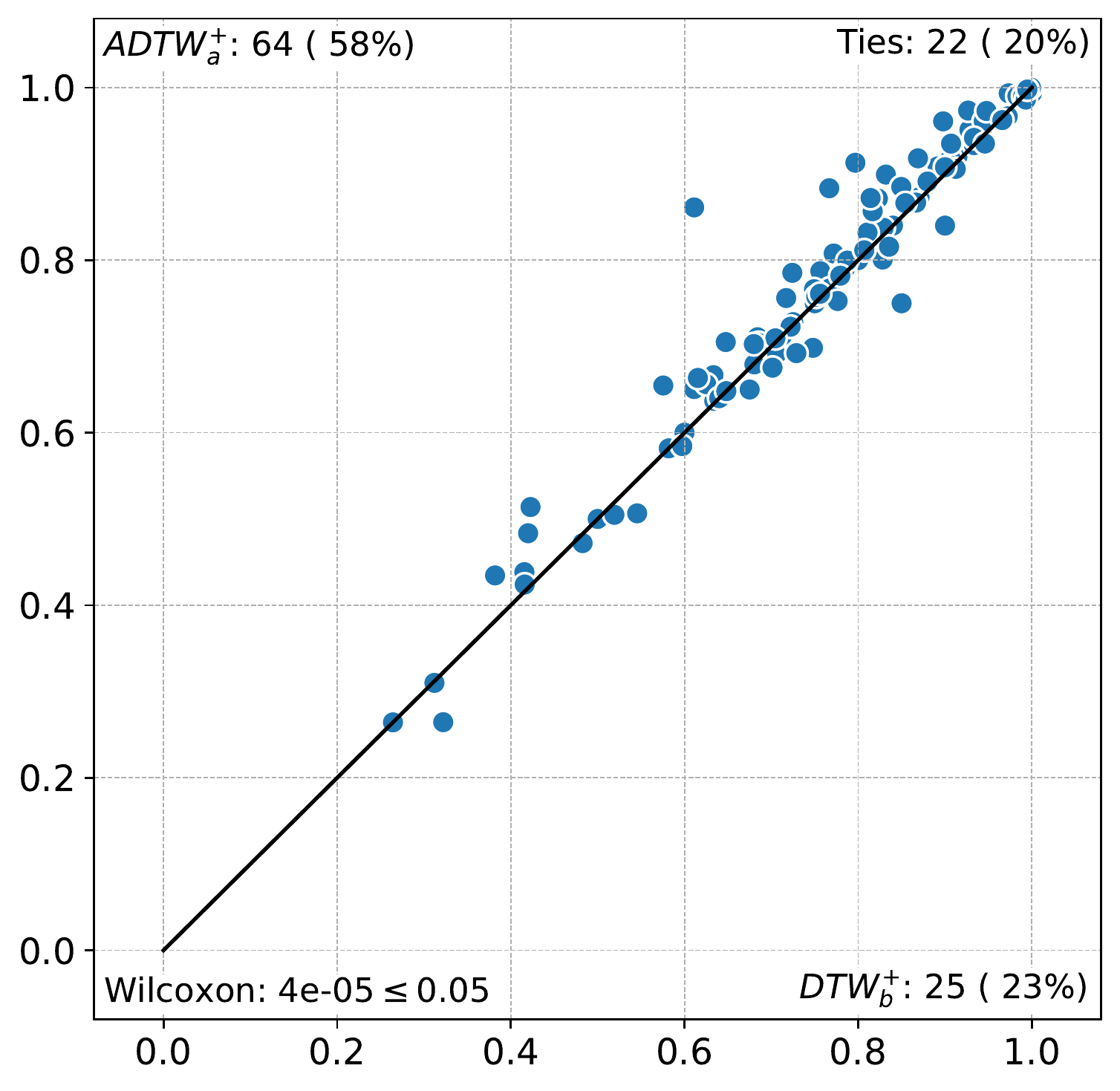}
        \caption{\label{fig:scp:adtw_dtw_b}$\ADTW^{+a}$ vs. $\DTW^{+b}$}
    \end{subfigure}
    \vspace*{-10pt}
    \caption{\label{fig:adtw_dtw}
                Accuracy scatter plot over the UCR archive comparing
                $\ADTW^{+a}$ against $\DTW^{+}$ tuned over $a$ and $b$.
                }
\end{figure}

From 
\cite{adtw}, $\ADTW^2$ is more accurate than $\DTW^2$.
Figure~\ref{fig:adtw_dtw} shows that \update{this remains the case for $\ADTW^{+a}$ and $\DTW^{+a}$}{$\ADTW^{+a}$ is also significantly more accurate than $\DTW^{+a}$.}
\update{More interestingly, it shows that $\ADTW^{+a}$ remains more accurate than $\DTW^{+b}$,
even though the latter benefits more from the larger set $b$ than the former.}{Interestingly, it also shows that $\ADTW^{+a}$ is also more accurate than $\DTW^{+b}$,
even though the latter benefits from the larger exponent set $b$.}

\subsection{Comparing PF vs $\PFvcfe$}

\begin{figure}
    \centering
    \includegraphics[width=0.49\textwidth]{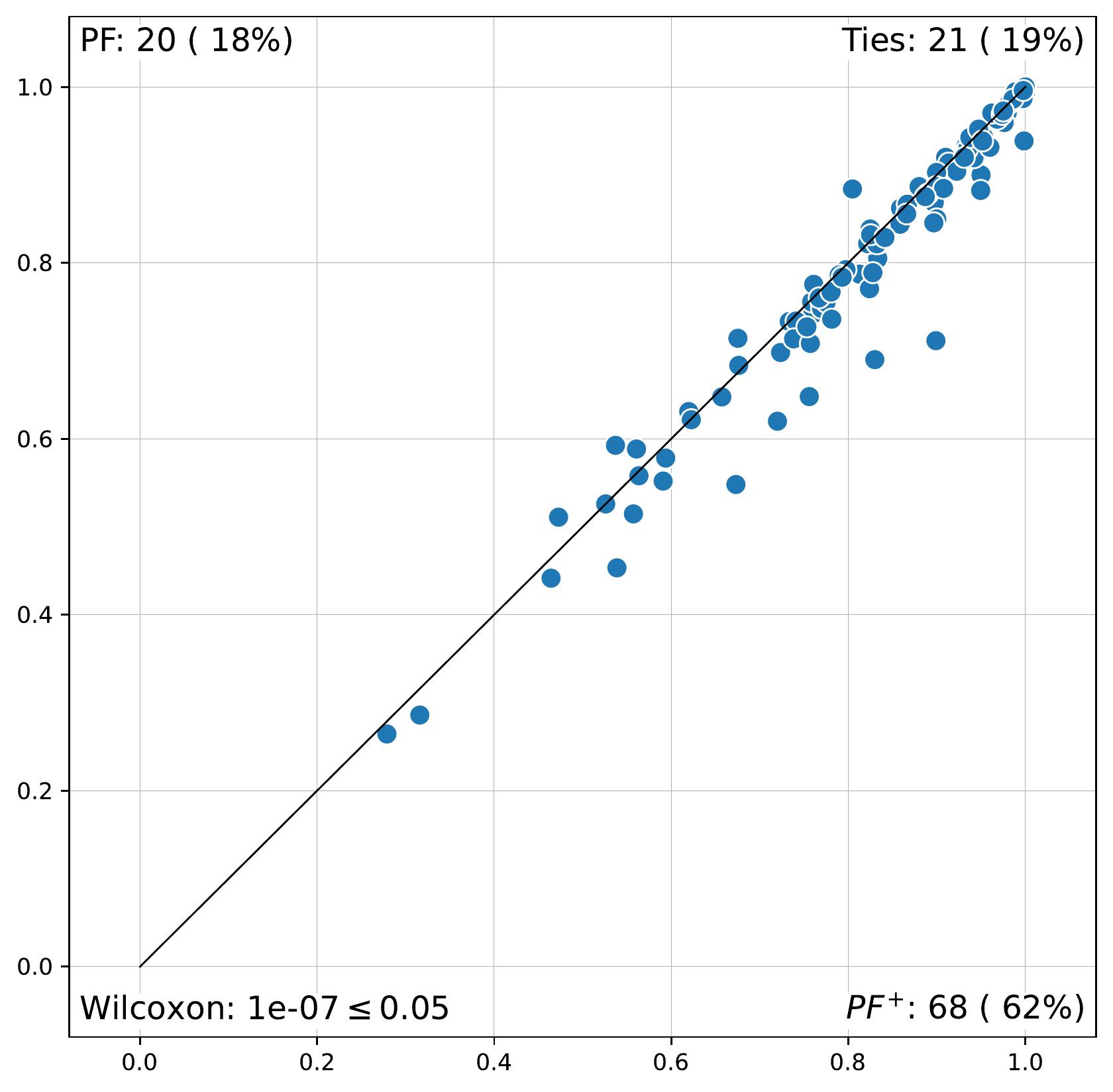}
    \vspace*{-10pt}
    \caption{\label{fig:pf_pfvcfe}
                Accuracy scatter plot over the UCR archive comparing
                the original Proximity Forest ($\PF$) 
                against
                Proximity Forest using $\lambda_{\gamma}$ for $\DTW$, and its variants ($\PFvcfe$).
                }
\end{figure}

Proximity Forest ($\PF$)~\citep{lucasProximityForestEffective2019} is an ensemble classifier relying on the
same 11 distances as the Elastic Ensemble (EE)~\citep{lines2015}, with the same parameter spaces.
Instead of using LOOCV to optimise each distance and ensemble their result,
PF builds trees of proximity classifiers, randomly choosing an exemplar, a distance and a parameter at each node.
This strategy makes it both more accurate and more efficient than EE \newtext{and the most accurate similarity-based time series classifier on the UCR benchmark}.

Proximity Forest and the Elastic Ensenble use the following distances:
the (squared) Euclidean distance ($\SQED$);
$\DTW$ with and without a window;
$\DDTW$ adding the derivative to $\DTW$~\citep{keoghDerivativeDynamicTime2001};
$\WDTW$~\citep{jeongWeightedDynamicTime2011a};
$\DWDTW$ adding the derivative to $\WDTW$;
$\LCSS$~\citep{hirschbergAlgorithmsLongestCommon1977};
$\ERP$~\citep{chenMarriageLpnormsEdit2004};
$\MSM$~\citep{stefanMoveSplitMergeMetricTime2013};
and $\TWE$~\citep{marteauTimeWarpEdit2009}.%

We define a new variant of Proximity Forest, $\PFvcfe$,
which differs only in replacing original cost functions for $\DTW$
and its variants by our proposed parameterized cost function.
We replace the cost function of $\DTW$ (with and without window), $\WDTW$, $\DDTW$, $\DWDTW$ and $\SQED$
by $\lambda_\gamma$, and randomly select $\gamma$ from the $a$ set at each node.
Note that the replacing the cost function of $\SQED$ in this manner makes it similar to a Minkowski distance.

We leave the tuning of other distances and their specific cost functions for future work.
This is not a technical limitation, but a theoretical one: we first have to ensure that such a change would
not break their properties.

The scatter plot presented in Figure~\ref{fig:pf_pfvcfe} shows that $\PFvcfe$ significantly outperforms $\PF$,
further demonstrating the value of extending the range of possible parameters to the cost function.
\begin{table}
    \caption{Six benchmark UCR datasets for which PF+ is more accurate than all four algorithms that have been identified as defining the current state of the art in TSC.}
    \label{tab:PFvsSOTA}
    \centering
\begin{tabular}{llllll}
\bf Dataset &
\bf $\PFvcfe$ &
\bf HC2 &
\bf TS-C &
\bf MR &
\bf IT \\
ArrowHead  &
0.8971 &
0.8629  &
0.8057  &
0.8629  &
0.8629   \\
Earthquakes & 
0.7698 &
0.7482 &
0.7482 &
0.7482 &
0.7410  \\
Lightning2 & 
0.8689 &
0.7869 &
0.8361 &
0.6885 &
0.8197 \\
SemgHandGenderCh2  &
0.9683 &
0.9567 &
0.9233 &
0.9583 &
0.8700 \\
SemgHandMovementCh2  & 
0.8800 &
0.8556 &
0.8778 &
0.7756 &
0.5689 \\
SemgHandSubjectCh2  &
0.9311 &
0.9022 &
0.9244 &
0.9244 &
0.7644 \\
\end{tabular}
\end{table}

\newtext{While similarity-based approaches no longer dominate performance across the majority of the UCR benchmark datasets, there remain some tasks for which similarity-based approaches still dominate.  Table~\ref{tab:PFvsSOTA} shows the accuracy of $\PFvcfe$ against four TSC algorithms that have been identified \citep{middlehurst2021hive} as defining the state of the art --- HIVE-COTE 2.0 \citep{middlehurst2021hive}, TS-CHIEF \citep{shifazTSCHIEFScalableAccurate2020}, MultiRocket \citep{tan2022multirocket} and InceptionTime \citep{fawaz2019inceptiontime}. This demonstrates that similarity-based methods remain an important part of the TSC toolkit.}

\section{Conclusion}\label{sec:conclusion}

$\DTW$ is a widely used time series distance measure.
It relies on a cost function to determine the relative weight to place on each difference between
values for a possible alignment between a value in one series to a value in another.
In this paper, we show that the choice of the cost function has substantial impact on nearest neighbor search tasks.
We also show that the utility of a specific cost function is task-dependent,
and hence that $\DTW$ can benefit from cost function tuning on a task to task basis.

We present a technique to tune the cost function by adjusting
the $\gamma$ exponent in a family of cost functions $\lambda_{\gamma}(a,b) = \abs{a-b}^{\gamma}$.
We introduced new time series distance measures utilizing this family of cost functions: $\DTW^+$ and $\ADTW^+$.
Our analysis shows that larger $\gamma$ exponents penalize alignments with large differences while smaller $\gamma$ exponents penalize alignments with smaller differences, allowing the focus to be tuned between small and large amplitude effects in the series.

We demonstrated the usefulness of this technique in both the nearest neighbor and Proximity Forest classifiers. \newtext{The new variant of Proximity Forest, $\PFvcfe$, establishes a new benchmark for similarity-based TSC, and dominates all of HiveCote2, TS-Chief, MultiRocket and InceptionTime on six of the UCR benchmark tasks, demonstrating that similarity-based methods remain a valuable alternative in some contexts.}


\newtext{We argue that cost function tuning can address noise through two mechanisms. Low exponents can exploit noise to penalize excessively long warping paths.  It appears that $\DTW$ benefits from this when windowing is not used.  High exponents direct focus to larger differences that are least affected by noise. It appears that $\ADTW$ benefits from this effect.}

We need to stress that we only experimented with one family of cost function,
on a limited set of exponents.
Even though we obtained satisfactory results,
we urge practitioners to apply expert knowledge when choosing their cost functions,
or a set of cost functions to select from.
Without such knowledge,
we suggest what seems to be a reasonable default set of choices for $\DTW^+$ and $\ADTW^+$,
significantly improving the accuracy over $\DTW$ and $\ADTW$.
We show that a \emph{denser} set does not substantially change the outcome,
while $\DTW$ may benefit from a \emph{larger} set that contains
more extreme values of $\gamma$ such as $0.2$ and $5$.

\newtext{A small number of exponents, specifically $0.5$, $1$ and $2$, lead themselves to much more efficient implementations than alternatives. It remains for future research to investigate the contexts in which the benefits of a wider range of exponents justify their computational costs.}

We expect our findings to be broadly applicable to time series nearest neighbor search tasks.
We believe that these finding also hold forth promise of benefit from
greater consideration of cost functions in the myriad of other applications of $\DTW$ and its variants.
 
\FloatBarrier

\section*{Acknowledgments}
\sloppy This work was supported by the Australian Research Council award DP210100072.
The authors would like to thank Professor Eamonn Keogh and his team at the University of California Riverside
(UCR) for providing the UCR Archive.

\section*{Declaration of interests}
The authors declare that they have no known competing financial interests or personal relationships
that could have appeared to influence the work reported in this paper.

\bibliographystyle{spbasic}
\bibliography{mybibfile}

\end{document}